\documentclass[final]{cvpr}

\usepackage{times}
\usepackage{epsfig}
\usepackage{graphicx}
\usepackage{amsmath}
\usepackage{amssymb}

\usepackage[pagebackref=true,breaklinks=true,colorlinks,bookmarks=false]{hyperref}

\usepackage{booktabs}       
\usepackage{caption,subcaption}
\usepackage{multirow}
\usepackage{listings}
\usepackage{wrapfig}
\usepackage{enumitem,kantlipsum}

\def\vs{\emph{vs.\ }}
\def\eg{\emph{e.g.\ }}
\def\ie{\emph{i.e.\ }}

\def\etal{\emph{et al.\ }}
\usepackage{xcolor}

\def\cvprPaperID{6195} 
\def\confYear{CVPR 2021}

\begin{document}

\title{The Heterogeneity Hypothesis: Finding Layer-Wise Differentiated Network Architectures}

\author{Yawei Li$^1$, Wen Li$^2$, Martin Danelljan$^1$, Kai Zhang$^1$, Shuhang Gu$^3$, Luc Van Gool$^{1, 4}$, Radu Timofte$^1$\\
$^1$Computer Vision Lab, ETH Z\"urich, $^2$UESTC, $^3$The University of Sydney, $^4$KU Leuven\\
{\tt\small \{yawei.li, martin.danelljan, kai.zhang, vangool, radu.timofte\}@vision.ee.ethz.ch}\\
{\tt\small \{liwenbnu,shuhanggu\}@gmail.com}
}


\maketitle
\pagestyle{empty}
\thispagestyle{empty}

\begin{abstract}
In this paper, we tackle the problem of convolutional neural network design. Instead of focusing on the design of the overall architecture, we investigate a design space that is usually overlooked, \ie adjusting the channel configurations of predefined networks. We find that this adjustment can be achieved by shrinking widened baseline networks and leads to superior performance. Based on that, we articulate the ``heterogeneity hypothesis'': with the same training protocol, there exists a layer-wise differentiated network architecture (LW-DNA) that can outperform the original network with regular channel configurations but with a lower level of model complexity.

The LW-DNA models are identified without extra computational cost or training time compared with the original network. This constraint leads to controlled experiments which direct the focus to the importance of layer-wise specific channel configurations. LW-DNA models come with advantages related to overfitting, \ie the relative relationship between model complexity and dataset size. Experiments are conducted on various networks and datasets for image classification, visual tracking and image restoration. The resultant LW-DNA models consistently outperform the baseline models. Code is available at \url{https://github.com/ofsoundof/Heterogeneity_Hypothesis.git}.
\end{abstract}

\section{Introduction}
\label{sec:introduction}

Since the advent of the deep learning era, convolutional neural network (CNN)~\cite{lecun1998gradient} design has replaced the role of feature design in various computer vision tasks.
Recently, neural network design has also evolved from manual design~\cite{simonyan2014very,he2016deep,huang2017densely} to neural architecture search (NAS)~\cite{liu2019darts,tan2019mnasnet} and semi-automation~\cite{yang2018netadapt,howard2019searching,radosavovic2020designing}. 
State-of-the-art network designs focus on discovering the overall network architecture with regularly repeated convolutional layers. 
This has been the golden standard of current CNN designs. For example, Ma~\etal mentioned that a network should have equal channel width~\cite{ma2018shufflenet}. But their analysis is limited to minimizing the memory access cost given the FLOPs for a single pointwise convolution. 

The motivation of this paper kind of contradicts the previous design heuristics. 
It investigates a design space that is usually overlooked and thus not fully explored, namely adjusting the layer-wise channel configurations. In this paper, the channel configuration of a network is defined as the vector that summarizes the output channels of the convolutional layers.
We try to answer three questions: 1) whether there exists a layer-wise differentiated network architecture (LW-DNA) that can outperform the original one; 2) if so, how to identify it efficiently; and 3) why it can beat the regular configuration.

\textbf{Question 1: The existence of LW-DNA.} To answer the first question, we formally articulate the following hypothesis.
\textit{\textbf{The Heterogeneity Hypothesis:} For a CNN, when trained with exactly the same training protocol (\eg number of epochs, batch size, learning rate schedule), there exists a layer-wise differentiated network architecture (LW-DNA) that can outperform the original network with regular layer-wise channel configurations but with a lower model complexity in term of FLOPs and parameters.}

To be specific, we aim at adjusting the numbers of channels of the convolutional layers in predefined CNNs. 
The other layer configurations such as kernel size and stride are not changed. 
Formally, consider an $L$-layer CNN $f(\mathbf{X};\boldmath{\Theta},\mathbf{c})$, where $\mathbf{c} = (c_1, c_2, \cdots, c_L)$ is the channel configuration of all of the convolutional layers, $\boldmath{\Theta}$ denotes the parameters in the network, and $\mathbf{X}$ is the input of the network.
The heterogeneity hypothesis implies that there should exist a new channel configuration $\mathbf{c}' = (c'_1, c'_2, \cdots, c'_L)$ such that the new architecture $f'(\mathbf{X};\boldmath{\Theta}',\mathbf{c}')$ performs no worse than the original one. 
After the adjustment, the channel configurations $c'_l$ could be either larger or smaller than the original $c_l$. 
We try to answer this question by empirical experiments.

\begin{figure}
    \centering
    \includegraphics[width=0.98\linewidth]{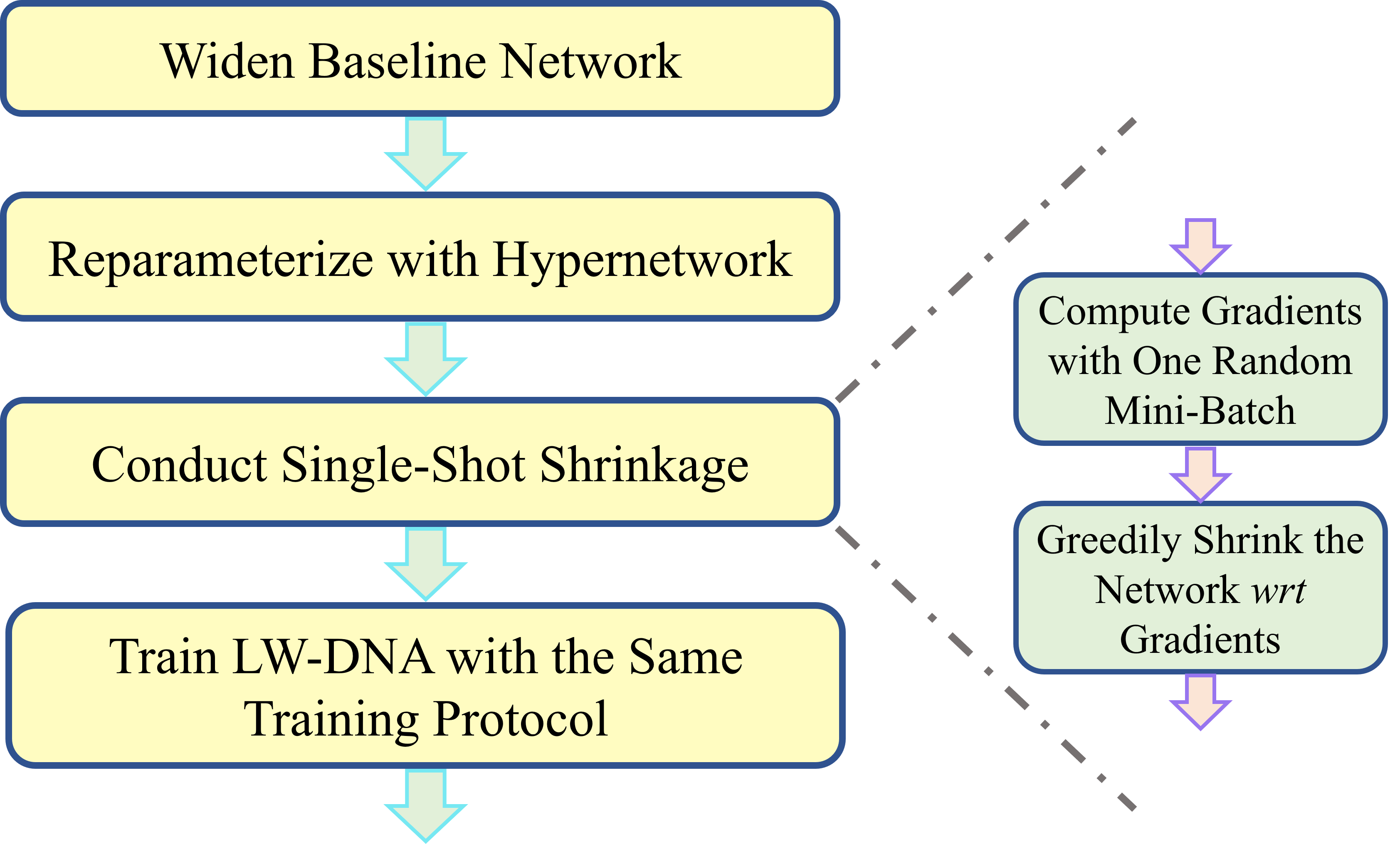}
    \caption{Pipeline of identifying LW-DNA models. Note that the single-shot shrinkage method only needs to run one random mini-batch. Then the network is shrunk after the single pass. Thus, almost no additional computational cost is introduced. This allows for fair comparison between the baseline model and the LW-DNA model.}
    \label{fig:lwdna_pipeline}
    \vspace{-0.6cm}
\end{figure}

\textbf{Question 2: How to identify an LW-DNA efficiently?} 
Note that the focus of this paper is solely the network architectures. The influence of factors other than network architecture such as the training protocol are excluded. This choice allows for controlled experiments and a fair comparison between the possibly existing LW-DNA models and the baseline models. But we are in turn faced with the following problem. 
\textit{\textbf{Problem Statement:} If the heterogeneity hypothesis is valid, how can we efficiently and reliably find an LW-DNA model for a CNN without additional computational cost and training time?}

To solve this problem, we are inspired by recent developments in network compression~\cite{lee2018snip,lee2019signal,li2020dhp}. The pipeline of identifying LW-DNA models is shown in Fig.~\ref{fig:lwdna_pipeline}. In short, the LW-DNA models are identified by the single-shot shrinking of a widened and reparameterized version of the baseline network. The details are given in Sec.~\ref{sec:methodology}

\textbf{Question 3: How to explain the benefits of LW-DNA?} 
As a matter of examples, we identify LW-DNA versions of various state-of-the-art networks for three vision tasks, incl. image classification~\cite{he2016deep,huang2017densely,howard2017mobilenets,sandler2018mobilenetv2,howard2019searching,tan2019mnasnet,tan2019efficientnet,radosavovic2020designing}, image restoration~\cite{ledig2016photo,lim2017enhanced,zhang2017beyond,ronneberger2015unet}, and visual tracking~\cite{bhat2019learning}. Interestingly, the identified LW-DNA models consistently outperform the baselines even with lower model complexities in terms of FLOPs and number of parameters. We try to explain this phenomenon from several perspectives.
\begin{enumerate}
    \item CNNs are redundant. So it is possible to find a layer-wise specific channel configuration comparable with the baseline under lower model complexity.
    \item As shown in Fig.~\ref{fig:channel_percentage}, some layers of the LW-DNA models have more channels than the baseline. Indeed, the lower layers tend to be strengthened with more channels. It might be those layers that play the essential role in improving the network accuracy. 
    \item The accuracy gain of the LW-DNA models might be related to overfitting by the baseline models. We derive this conjecture from several observations. \textbf{I.} By comparing the training and testing curves of an LW-DNA model and its baseline in Fig.~\ref{fig:log}, we find that towards the end of the training, the identified LW-DNA model shows a higher training error but a lower testing error, \ie improved generalization. This phenomenon is consistent across different datasets. This also matches the observations from the pioneering unstructured pruning, like a brain surgeon trying to boost network generalization after brain damage~\cite{lecun1990optimal,hassibi1993second}. \textbf{II.} The accuracy gain of an LW-DNA model is larger for smaller datasets (\ie Tiny-ImageNet) that are easier to get overfitted to, compared with larger datasets (\ie ImageNet). \textbf{III.} On the same dataset (\ie ImageNet), it is easier to identify an LW-DNA model version for larger networks (\ie ResNet50) than for smaller networks (\ie MobileNetV3). 
\end{enumerate}

The contributions of this paper can be summarized as follows. \textbf{First}, it demonstrates the possibility of identifying a superior version of a network by only adjusting the channel configuration of the network. This could be used as a post-searching mechanism complementary to semi- or fully automated neural architecture search. \textbf{Secondly}, a method that can identify LW-DNA models almost without additional computational cost and training time is proposed. This method only needs the computation of one random batch.
\textbf{Thirdly}, the possible reason for the improved performance of an LW-DNA is explained by observing the experimental results.

\section{Related Work}

\textbf{The lottery ticket hypothesis (LTH).} The heterogeneity hypothesis is reminiscent of the LTH~\cite{frankle2018lottery}, which addresses the existence of sparse subnetworks that can match the test accuracy of randomly-initialized dense networks. The winning ticket is identified by greedily pruning single elements of weight parameters with smallest magnitude. Following works try to extend~\cite{ramanujan2020s}, theoretically prove~\cite{malach2020proving}, understand~\cite{zhou2019deconstructing}, and improve the training process~\cite{renda2020comparing} of LTH. The unstructured pruning breaks the dynamical isometry in the network~\cite{lee2019signal}. The core problem is the trainability of the sparse subnetworks and the gradient flow in the subnetworks~\cite{lee2019signal}. In contrast, the heterogeneity hypothesis focuses on adjusting the channel configuration of the network. Since the weight elements of an entire channel are pruned together, there is no irregular kernel in the pruned network. Gradient flow is no longer a problem in this scenario. 

\textbf{NAS.} NAS automatizes neural network design by searching in the design space~\cite{zoph2017nas,liu2018nas,pham2018efficient}. Earlier works consume lots of computation~\cite{zoph2017nas,liu2018nas}. Recent developments accelerate the searching procedure by introducing differentiability into searching~\cite{liu2019darts}. After the searching stage, the derived cells are repeated to construct the final network. Thus, the final network still has a regular architecture. In this paper, we try to adjust the channel configuration of the network, which can be regarded as a method complementary to NAS. Recent works try to push the frontier of NAS research by either redesigning the search space or proposing a more efficient search method~\cite{xu2019pc,ru2020neural,guo2020single,chen2019progressive}. 

\textbf{Hypernetworks.} Hypernetworks are actually a kind of reparameterization of the backbone network~\cite{ha2017hypernetworks}. Hypernetworks generate the weight parameters of the backbone network. The input of hypernetworks can be either static or dependent on the feature maps of the backbone network. In this sense, hypernetworks fall under the paradigm of meta learning. Recent developments bring hypernetworks to network compression~\cite{liu2019metapruning,li2020dhp}. Earlier hypernetwork designs are just a stack of two linear layers. Thus, the outputs are fixed, which should be cropped before being used as weights of the backbone network. The recent hypernetworks~\cite{li2020dhp} can adapt the outputs according to the length of the input latent vectors. This design naturally suits the task of network compression. This is one of the reasons why we select hypernetworks as our shrinkage agent.

\textbf{Network shrinkage.} Network shrinkage removes unimportant weight parameters in the network~\cite{lecun1990optimal,hassibi1993second,han2015deep,li2017pruning,li2020group,lee2018snip,li2020dhp}. 
Since we want to purely investigate the importance of the architecture of the identified network, the other factors such as training protocol should be excluded. The network shrinkage procedure should also be simplified as much as possible. Inspired by \cite{lee2018snip, lee2019signal}, the widened network is shrunk at initialization according to gradients. The network shrinkage procedure only needs one random batch.

\textbf{Difference from network compression works.}
This paper is different from the previous network compression works in the following aspects.
\textbf{Aim.} The aim of this paper is a a proof of a concept that it is possible to benefit better from the computation and parameter budget by optimizing the architecture of the network. The identified LW-DNA model of a predefined network has improved accuracy and slightly reduced model complexity. 
Previous network compression works aim at improving the efficiency of networks. Accuracy drop is inevitable for the compact networks. 
\textbf{Method.} The single-shot method in~\cite{lee2018snip} for unstructured pruning is transferred to network shrinkage by its collaboration with hypernetworks. There is no computational overhead for the network shrinkage method used in this paper.
\textbf{Interpretation.} This paper tries to interpret where the benefit of the slightly reduced models comes from, which is not done by recent works.

\section{Preliminaries}
\label{sec:preliminary}

\subsection{Hints from network compression} 
Recent network compression methods shed light on the existence of advantageous layer-wise specific networks~\cite{liu2019metapruning,li2020dhp,ding2020lossless}. 
Those methods can result in shrunk networks with layer-wise specific channel configurations.
Some works~\cite{liu2019metapruning} report accuracy gains of the pruned network over the width-scaled versions of ResNet and MobileNets~\cite{he2016deep,howard2017mobilenets,sandler2018mobilenetv2}.
Yet, since the advantageous networks are identified in a network compression sense, thus with an accuracy drop compared with the uncompressed network, it still remains unknown whether there exists a layer-wise specific network that can compete with the original one. 
A recent work~\cite{li2020dhp} reports an accuracy gain over uncompressed MobileNets on Tiny-ImageNet. Yet, further investigations on larger datasets are not conducted.
Moreover, the compact networks are usually derived with training protocols different from those used for the baseline network, \eg additional searching stage, larger batch size, or prolonged fine-tuning stage. It remains unknown how the layer-wise specific channel configurations benefit the network. 

\begin{figure}
    \centering
    \includegraphics[width=0.8\linewidth]{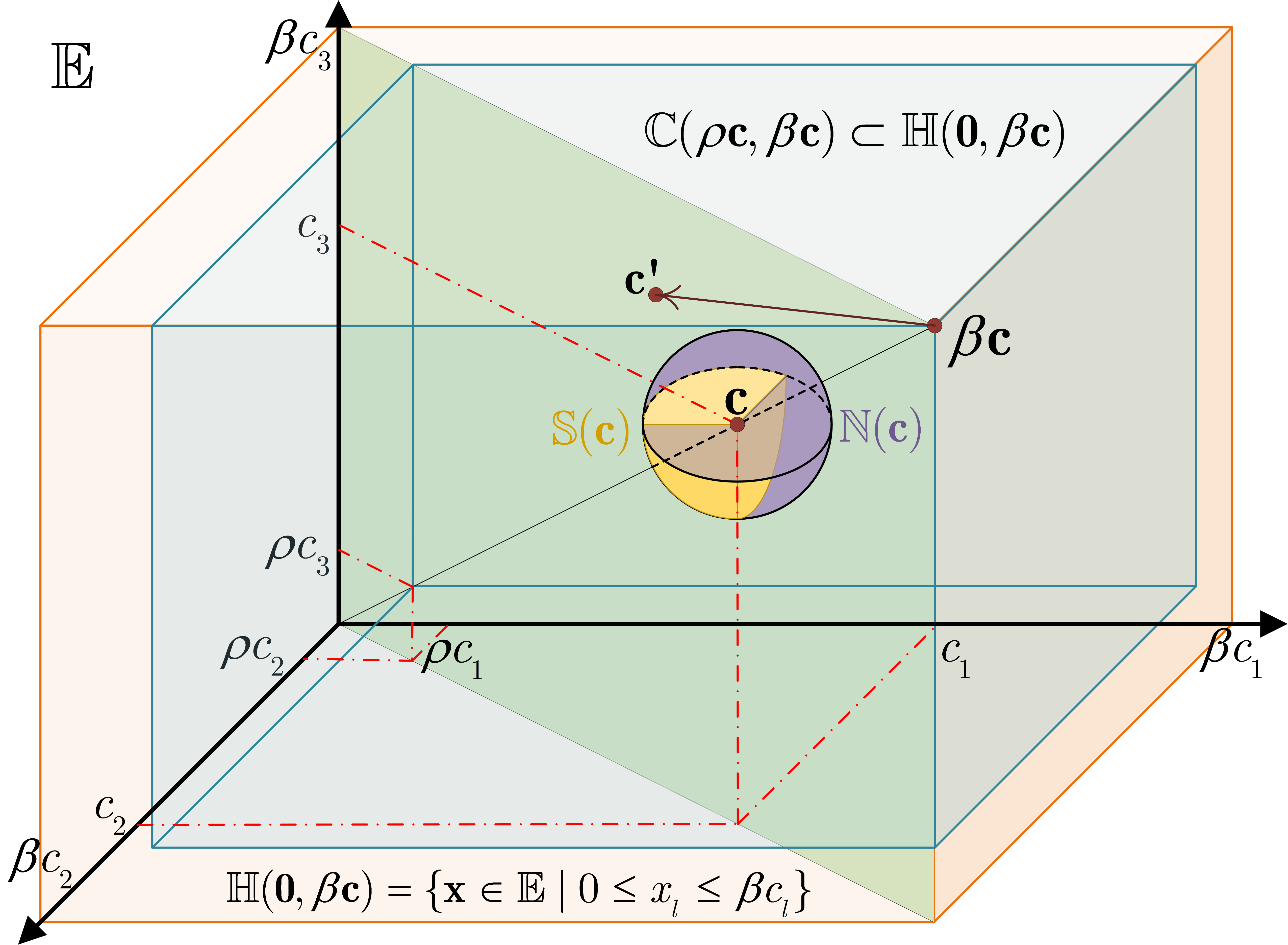}
    \caption{Illustration of the configuration space. The proposed method identifies layer-specific channel configurations within the enlarged and constrained subspace $\mathbb{C}(\rho \mathbf{c}, \beta \mathbf{c})$. Compared with searching within the constrained neighborhood $\mathbb{S}(\mathbf{c})$ of $\mathbf{c}$, the enlarged configuration space makes it possible to develop a straightforward shrinkage criterion.}
    \label{fig:configuration_space}
    \vspace{-0.6cm}
\end{figure}

\subsection{Notations and definitions}

\textbf{Notation}. In this paper, bold lowercase letters such as $\mathbf{c}$, $\mathbf{x}$, and $\mathbf{z}$ are used to denote vectors while bold capital letters such as $\mathbf{O}$, $\mathbf{Z}$, $\mathbf{W}$ are used to denote matrices and higher dimensional tensors. The vectors, matrices, and higher dimensional tensors are indexed by subscripts. Greek letters such as $\alpha$, $\beta$ denote constant scalars. The configuration vector and configuration space are formally defined as follows.

\textbf{Definition 1} (Channel configuration vector). Consider an $L$-layer CNN. The channel configuration vector of the network is defined as an $L$-dimensional vector that summarizes the number of output channels of the network, \ie
\begin{equation}
    \mathbf{c} = (c_1, c_2, \cdots, c_L),
\end{equation}
where $c_l$ denotes the number of output channels in the $l$-th layer.

\textbf{Definition 2} (Configuration space). The configuration space $\mathbb{E}$ is a subspace of Euclidean space that contains the allowable channel configuration vectors. (See Fig.~\ref{fig:configuration_space} for one example of the configuration space.) 

The dimension of the configuration vectors depends on the number of convolutional layers in the network. Take VGG11 for example. The configuration vector is an 8-dimensional vector, \ie,
\begin{equation}
    \mathbf{c}_{vgg} = (64, 128, 256, 256, 512, 512, 512, 512)\,.
\end{equation}
As in this example, the configuration vector is regular and its elements are dependent on each other in the sense that most of them are repeated. 
For image classification networks, the golden standard is to repeat building blocks with the same configuration up to the point where the spatial dimension of the feature map gets reduced. 
Some efficient designs for mobile devices introduce a width multiplier $\alpha$ to adapt to constrained resource requirements, which results in a scaled configuration vector, \ie,
\begin{equation}
    \mathbf{b} = (\alpha c_1, \alpha c_2, \cdots, \alpha c_L),\, \alpha < 1\,. 
\end{equation}

\subsection{Problem formulation and recast}
\label{subsec:formulation}

Since the configuration vector is manually fixed, it is not guaranteed to be optimal. In this paper, we explore the corresponding configuration design space.
The aim is to demonstrate that there is an irregular configuration vector $\mathbf{c}'$ that can compete with the original, while offering reduced model complexity. 
To achieve that, we propose an algorithm which can adjust (increase or decrease) the elements of the configuration vector $\mathbf{c}$ while controlling the model complexity. 
As shown in Fig.~\ref{fig:configuration_space}, such an adjustment procedure best searches in the neighborhood of the vector, \ie
\begin{equation}
    \mathbb{N}(\mathbf{c}) \subset \mathbb{E}\,.
\end{equation}
After the adjustment, an element of the configuration vector $\mathbf{c}$ can be either increased or decreased, which corresponds to growing or shrinking the $l$-th layer of the network.
Shrinkage criteria can be defined on the existing network and network shrinkage algorithm could applied.
The limitation of a shrinkage algorithm on the original network is that it can only explore a subspace of the neighborhood, \ie
\begin{equation}
    {\mathbb{S}}(\mathbf{c}) = \{\mathbf{x}\in{\mathbb{N}}(\mathbf{c})|x_l \leq c_l\} \subset \mathbb{N}(\mathbf{c})\,.
\end{equation}
But we do not want to be restricted to shrinkage only. Instead, it is desirable to do both network shrinkage and growth at the same time for the configuration vector adjustment.

We circumvent this problem by recasting it as a shrinkage problem in a larger configuration space which is obtained by widening the width of the network with a width multiplier $\beta > 1$. 
The new searching space $\mathbb{H}$ is a hyperrectangle delimited by the zero vector $\mathbf{0}$ and the up-scaled configuration vector $\beta \mathbf{c}$ in the high-dimensional space, \ie 
\begin{equation}
    \mathbb{H}(\mathbf{0}, \beta \mathbf{c}) = \{\mathbf{x} \in \mathbb{E}|0 \leq x_l \leq \beta c_l\} \subset \mathbb{E}\,.
    \label{eqn:larger_configuration_space}
\end{equation}
The searching algorithm then starts from the up-scaled vector $\beta \mathbf{c}$ and reduces the value of its $l$-th element greedily according to the significance of the channels in the corresponding convolutional layer.

\section{Methodology}
\label{sec:methodology}

After introducing the preliminaries and the designing considerations in the last section, the algorithm used to identify LW-DNA models is explained in this section. 
The pipeline is already shown in Fig.~\ref{fig:lwdna_pipeline}. 
The identifying procedure proceeds as follows. 
1) Reparameterize the widened baseline network with hypernetworks. The outputs of the hypernetworks act as the weight parameters of the baseline network. The inputs of the hypernetwork serve as the handle to shrink the network. 
2) Compute the gradients of the hypernetwork input, \ie the latent vectors, with one random batch. 
3) Sparsify the latent vectors greedily according to the magnitude of their gradients.
4) Compute the weight parameters with the sparsified latent vectors. 
5) Train the resultant network from scratch with the same training protocol as the baseline network.
And in the following, we explain some of the key steps in detail.

\subsection{Reparameterizing with hypernetworks}
\label{subsec:reparameterization}

The network shrinkage method is explained in this section. Instead of directly shrinking the baseline network, we first widen it and reparameterize it with hypernetworks~\cite{liu2019metapruning,li2020dhp}. The reparameterization is adopted based on the following considerations.
The hypernetworks bring the shrinkage problem into a latent space. Removing a channel is equivalent to deleting a single element of the latent vector, which converts the problem of dealing with elements in the whole channel to an easier one of dealing with a single element in the latent vector.  
In addition, it provides a straightforward extension of single-shot shrinkage~\cite{lee2018snip} to channel pruning (See Subsec~\ref{subsec:single_shot_shrinkage}). And single-shot shrinkage is the core of avoiding additional computational cost when identifying LW-DNA models. 
The latent vector sharing mechanism in the hypernetworks also makes it possible to deal with various state-of-the-art networks. 

Consider the $L$-layer CNN that is brought to the larger configuration space $\mathbb{H}(\mathbf{0}, \beta \mathbf{c})$ as in Eqn.~\eqref{eqn:larger_configuration_space}.
The weight parameter of the $l$-th convolutional layer of the CNN has the dimension of $\beta c_{l} \times \beta c_{l-1} \times w \times h$, where $\beta c_{l}$, $\beta c_{l-1}$, and $w \times h$ denotes the output channel, input channel, and kernel size of the layer.
For the simplicity of notation, let $n = \beta c_{l}$ and $c = \beta c_{l-1}$ in the following.

\begin{figure}[!t]
    \begin{center}
       \includegraphics[width=1.0\linewidth]{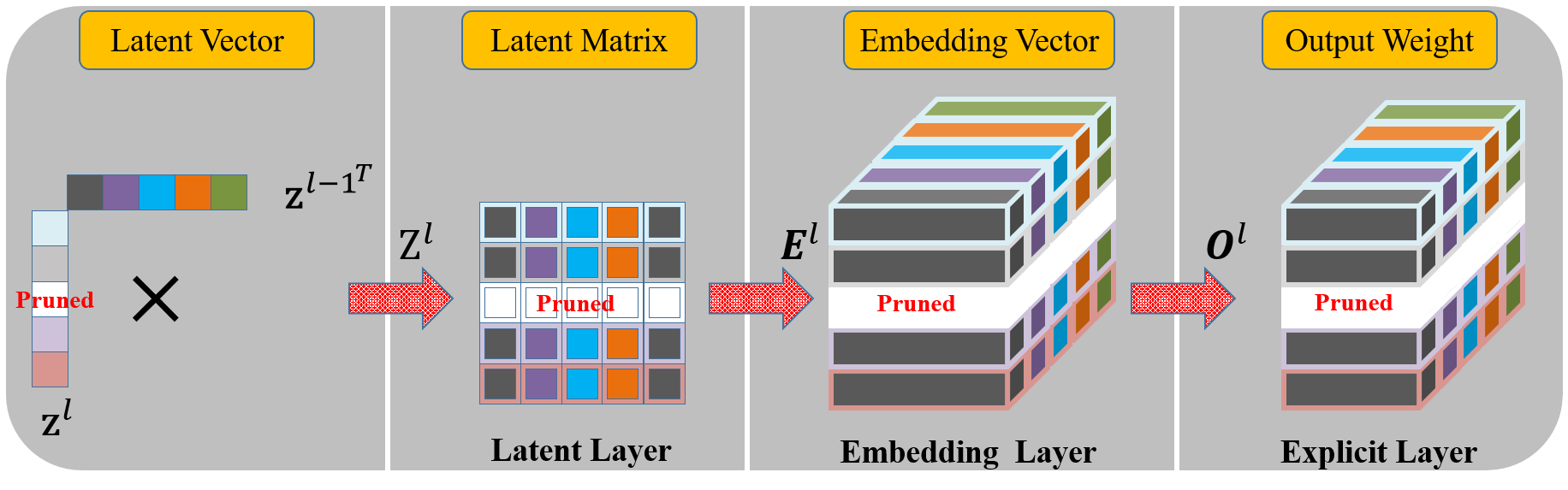}
    \end{center}
    \vspace{-12pt}
    \caption{Illustration of the operations in the hypernetworks.}
    \vspace{-10pt}
\label{fig:hypernetwork}
\end{figure}

To reparameterize the $l$-th layer of the network, a latent vector $\mathbf{z}^l \in \mathbb{R}^n$ is first attached to it.
The latent vector controls the output channel of the layer and removing an element of the latent vector is equivalent to deleting an output channel of the layer. 
Since the output channel of the current layer acts as the input channel of the next layer, the latent vectors are shared between consecutive layers. 
Then as shown in Fig.~\ref{fig:hypernetwork}, the hypernetwork takes as input the latent vector of the previous layer and the current layer. It computes a latent matrix, \ie $\mathbf{Z}^l = {\mathbf{z}^{l}} \cdot {\mathbf{z}^{l-1}}^{T}$, which forms a grid used for network shrinkage. Every element of the latent matrix is transformed to a vector by two consecutive linear operations, \ie 
\begin{equation}
    \mathbf{O}^l_{i,j} =  \mathbf{W}^l_2 \cdot (\mathbf{Z}^{l}_{i,j} \mathbf{W}^l_1)\,, 
    \label{eqn:explicit}
\end{equation}
where $\mathbf{W}^l_1 \in \mathbb{R}^{m \times 1}$ transforms the elements into a higher-dimensional embedding space and $\mathbf{W}^l_2 \in \mathbb{R}^{wh \times m}$ converts the tensors vectors in the embedding space to the output weight. 
Note that $\mathbf{W}^l_1$ and $\mathbf{W}^l_2$ are unique for each element $\mathbf{Z}^{l}_{i,j}$ and for the simplicity of notation the subscript ${i,j}$ is omitted.
The output could be assembled into a 3D tensor $\mathbf{O}^l \in \mathbb{R}^{n \times c \times wh}$ which can be used as the weight parameter of the convolutional layer.
The latent matrix acts as a handle to shrink the layer. As shown in Fig.~\ref{fig:hypernetwork}, if a single element of the latent vector ${\mathbf{z}^l}$ is nullified, the entire row of the latent matrix is masked out. As a result, the corresponding output channel is removed.
Similarly, removing an element of ${\mathbf{z}^{l-1}}$ corresponds to removing the output channel of the $l-1$-th layer and the input channel of the $l$-th layer.
Details of handling different layers such as depth-wise convolution and batch normalization are given in the supplementary.

\subsection{Single-shot shrinkage}
\label{subsec:single_shot_shrinkage}

After reparameterizing the network with hypernetworks, the parameters in the network are first randomly initialized~\cite{chang2020principled}. Then the single-shot shrinkage method is used to adjust the width of the network. 

Consider a single mini-batch $\{\mathbf{X}_i, \mathbf{Y}_i\}$ from the dataset. The output of the network is computed as
\begin{equation}
    \mathbf{\hat{Y}}_i = f(\mathbf{X}_i; {\boldmath{\Theta}}, \mathbf{z})\,,
\end{equation}
where $\mathbf{z}$ denotes the latent vectors and $\boldmath{\Theta}$ is the parameter set that contains $\mathbf{W}_1$ and $\mathbf{W}_2$.
The loss is computed as
\begin{equation}
    \mathcal{L} = \mathcal{L}\left(\mathbf{Y}_i, f(\mathbf{X}_i; \boldmath{\Theta}, \mathbf{z})\right)\,.
\end{equation}
Then the gradients of the loss function with respect to the latent vectors are computed as 
\begin{equation}
    \nabla\mathcal{L} = \frac{\partial \mathcal{L}\left(\mathbf{Y}_i, f(\mathbf{X}_i; \boldmath{\Theta}, \mathbf{z})\right)}{\partial \mathbf{z}}\,.
\end{equation}
The magnitude of the gradients is used as the criterion to sparsify the latent vectors. 
The elements whose gradient magnitude is smaller than a threshold are removed.
The threshold is determined by a binary search algorithm, which allows the resultant network to reach a predefined FLOP target.
The resultant network is the final LW-DNA model and is trained from scratch with the same training protocol as the baseline model. 

The single-shot shrinkage method is inspired by single-shot pruning of weight elements~\cite{lee2018snip}.
But the original method is single element oriented. It removes single weight parameters in the network and results in unstructured kernels. 
It remains to be explored how to transform the single-shot method to network shrinkage. 
The hypernetworks provide such a connection. 
By resorting to hypernetworks, the shrinkage is conducted on the latent space whose elements correspond to channels in the network and serve as the agent for shrinkage. 
Deleting an element of the latent vector is equivalent to remove a channel in the network.
Thus, sparsifying the latent vectors according to their gradients is a natural transferring of the single-shot method in~\cite{lee2018snip}.

\begin{table*}[!t]
    \small
    \begin{center}
        \begin{tabular}{c|c|c|c|c|c}
            \toprule
             
             Dataset & Network  & Method & Top-1 Error (\%) & FLOPs [G] / Ratio (\%) & Params [M] / Ratio (\%) \\ \midrule
            
            \multirow{9}{*}{ImageNet~\cite{deng2009imagenet}} & \multirow{5}{*}{ResNet50~\cite{he2016deep}}
            &Baseline	&23.28	&4.1177	/ 100.0	&25.557	/ 100.0	\\
            &&MutualNet~\cite{yang2020mutualnet} &21.40 & 4.1177 / 100.0 &25.557 / 100.0\\
            &&LW-DNA	&23.00	&3.7307	/ 90.60	&23.741	/ 92.90	\\ 
            &&MetaPruning~\cite{liu2019metapruning} & 23.80 & 3.0000 / 72.86 & -- \\
            &&AutoSlim~\cite{yu2019autoslim}  &24.00 & 3.0000 / 72.86 & 23.100 / 90.39 \\ \cline{2-6}
            &\multirow{2}{*}{\shortstack{RegNet~\cite{radosavovic2020designing} \\ X-4.0GF}} 
            &Baseline	&23.05	&4.0005	/ 100.0	&22.118	/ 100.0	\\
            &&LW-DNA	&22.74	&3.8199	/ 95.49	&15.285	/ 69.10	\\ \cline{2-6}
            &\multirow{2}{*}{\shortstack{MobileNetV3 small~\cite{howard2019searching}}} 
            &Baseline	&34.91	&0.0612 / 100.0	&3.108	/ 100.0	\\
            &&LW-DNA	&34.84	&0.0605	/ 98.86 &3.049	/ 98.11	\\ \midrule
            
            \multirow{12}{*}{Tiny-ImageNet} & \multirow{4}{*}{MobileNetV1~\cite{howard2017mobilenets}} 
            &Baseline	&51.87	&0.0478	/ 100.0	&3.412	/ 100.0	\\
            &&Baseline KD	&48.00	&0.0478	/ 100.0	&3.412	/ 100.0	\\
            &&DHP KD	&46.70	&0.0474	/ 99.16	&2.267	/ 66.43	\\
            &&LW-DNA	&46.44	&0.0460	/ 96.23	&1.265	/ 37.08	\\ \cline{2-6}
        	&\multirow{4}{*}{MobileNetV2~\cite{sandler2018mobilenetv2}} 	
            &Baseline	&44.38	&0.0930	/ 100.0	&2.480	/ 100.0	\\
            &&Baseline KD	&41.25	&0.0930	/ 100.0	&2.480	/ 100.0	\\
            &&DHP KD	&41.06	&0.0896	/ 96.34	&2.662	/ 107.34	\\
            &&LW-DNA	&40.74	&0.0872	/ 93.76	&2.230	/ 89.90	\\ \cline{2-6}
            
            &\multirow{4}{*}{\shortstack{MobileNetV3 small~\cite{howard2019searching}}}
            &Baseline	&47.55	&0.0207	/ 100.0	&2.083	/ 100.0	\\
            &&Baseline KD	&41.52	&0.0207	/ 100.0	&2.083	/ 100.0	\\
            &&DHP KD	&41.46	&0.0192	/ 92.75	&1.078	/ 51.76	\\
            &&LW-DNA	&41.35	&0.0178	/ 85.99	&1.799	/ 86.36	\\ \cline{2-6}
            
            &\multirow{4}{*}{MnasNet~\cite{tan2019mnasnet}} 					
            &Baseline	&51.79	&0.0271	/ 100.0	&3.359	/ 100.0	\\
            &&Baseline KD	&48.17	&0.0271	/ 100.0	&3.359	/ 100.0	\\
            &&DHP KD	&48.10	&0.0264	/ 97.42	&2.512	/ 74.79	\\
            &&LW-DNA	&46.85	&0.0250	/ 92.25	&1.258	/ 37.45	\\ \midrule
            
            \multirow{9}{*}{CIFAR100} 

            &\multirow{2}{*}{\shortstack{RegNet~\cite{radosavovic2020designing} \\ Y-400MF}}
            &Baseline	&21.65	&0.4585	/ 100.0	&3.947	/ 100.0	\\
            &&LW-DNA	&18.65	&0.4468	/ 97.45	&2.466	/ 62.48	\\ \cline{2-6}

            &\multirow{2}{*}{\shortstack{RegNet~\cite{radosavovic2020designing} \\ X-400MF}}
            &Baseline	&21.75	&0.4698	/ 100.0	&4.810	/ 100.0	\\
            &&LW-DNA	&18.81	&0.4610	/ 98.13	&4.404	/ 91.56	\\ \cline{2-6}
            &\multirow{2}{*}{EfficientNet~\cite{tan2019efficientnet}}					
            &Baseline	&20.74	&0.4161	/ 100.0	&4.136	/ 100.0	\\
            &&LW-DNA	&19.54	&0.3850	/ 92.53	&2.121	/ 51.28	\\ \cline{2-6}
            &\multirow{2}{*}{DenseNet40~\cite{huang2017densely}}					
            &Baseline	&26.00	&0.2901	/ 100.0	&1.100	/ 100.0	\\
            &&LW-DNA	&22.46	&0.2638	/ 90.93	&1.016	/ 92.35	\\ \midrule
            
            \multirow{2}{*}{CIFAR10~\cite{krizhevsky2009learning}} & \multirow{2}{*}{DenseNet40~\cite{huang2017densely}}
            &Baseline	&5.50	&0.2901	/ 100.0	&1.059	/ 100.0	\\
            &&LW-DNA	&4.87	&0.2632	/ 90.73	&0.963	/ 90.87	\\ \cline{2-6}
            &\multirow{2}{*}{ResNet56~\cite{he2016deep}}				
            &Baseline	&5.74	&0.1274	/ 100.0	&0.856	/ 100.0	\\
            &&LW-DNA	&5.49	&0.1262	/ 99.06	&0.536	/ 62.62	\\ \midrule

        \end{tabular}
    \end{center}
    \vspace{-0.4cm}
    \caption{Image classification results.
    Baseline and Baseline KD denote the original network trained without and with knowledge distillation respectively.}
    \label{tbl:results_classification}
    \vspace{-0.6cm}
\end{table*}

\subsection{Constraining model complexity}
\label{subsec:model_complexity}

Model complexity is measured in terms of FLOP and parameter count. 
The target is to find a model that has both fewer FLOPs and parameters while achieving improved accuracy. 
Yet, the two metrics are not always consistent with each other. 
For example, when the FLOPs target is set, a parameter over-pruned model might be observed in some of the experiments, which could lead to inferior performance. 
Thus, a new hyper-parameter $\rho$ is introduced which controls the minimum percentage of remaining channels in convolutional layers. 
In this way, the search space $\mathbb{C}(\rho \mathbf{c}, \beta \mathbf{c})$ is a confined subspace of the original search space $\mathbb{H}(\mathbf{0}, \beta \mathbf{c})$, \ie
\begin{equation}
    \mathbb{C}(\rho \mathbf{c}, \beta \mathbf{c}) = \{\mathbf{x} \in \mathbb{E}|\rho c_l \leq x_l \leq \beta c_l\} \subset \mathbb{H}(\mathbf{0}, \beta \mathbf{c})\,.
\end{equation}
A similar hyper-parameter $\tau$ is introduced for the final linear layers  of image classification networks.
The hyper-parameters $\rho$ and $\tau$ are termed convolutional percentage and linear percentage in this paper, respectively.  
During the pruning, the FLOP budget is fixed. By tuning the hyper-parameters $\rho$ and $\tau$, the algorithm is able to find networks with the same FLOPs but varying parameter budgets.

\section{Experimental Results}
\label{sec:results}

\begin{figure*}[!hbt]
\begin{minipage}[c]{0.33\textwidth}
  \vspace*{\fill}
  \centering
  \includegraphics[width=1\linewidth]{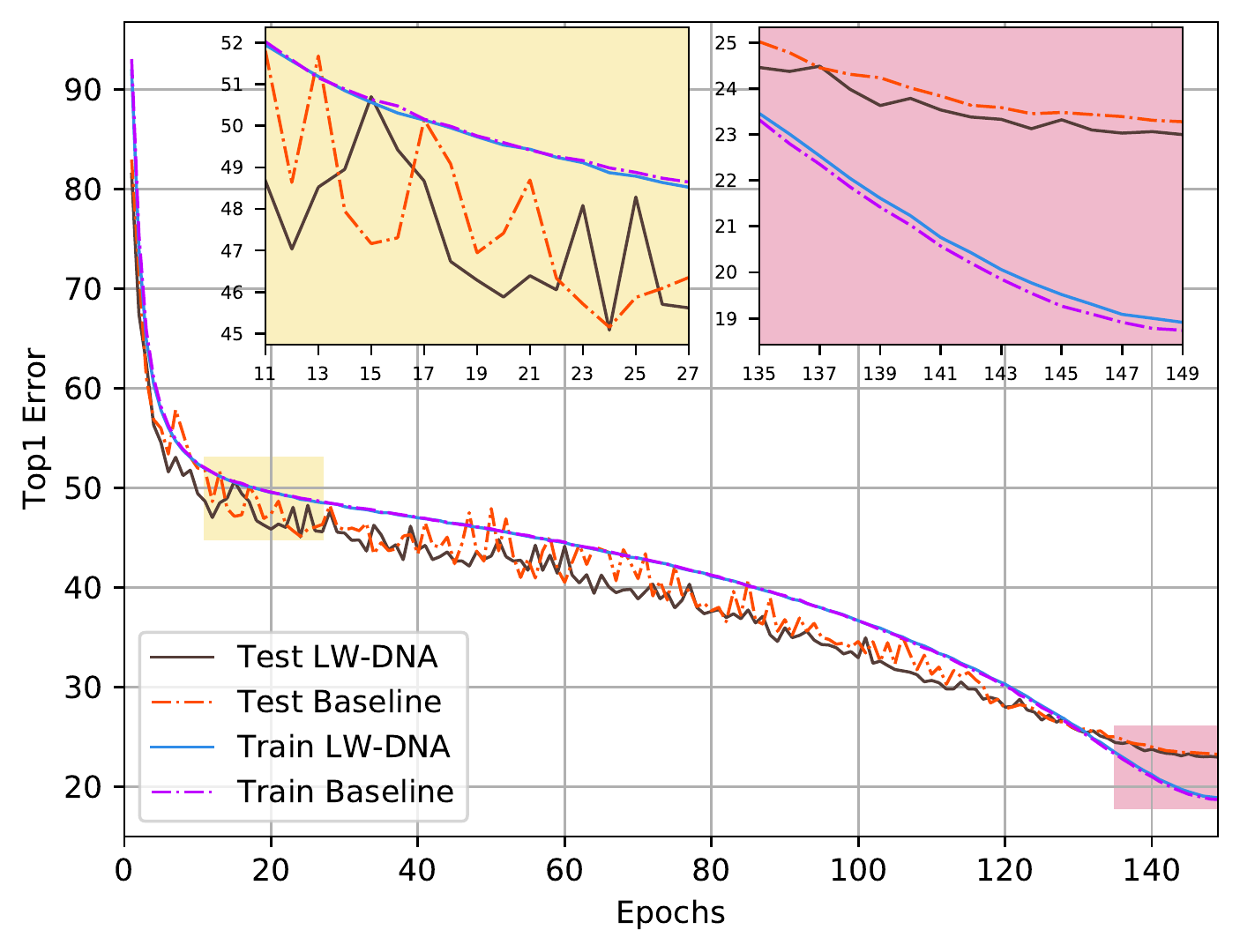}
  \vspace{-0.6cm}
  \subcaption{ResNet50, ImageNet.}
  \label{fig:log_resnet50}
\end{minipage}%
\begin{minipage}[c]{0.33\textwidth}
  \vspace*{\fill}
  \centering
  \includegraphics[width=1\linewidth]{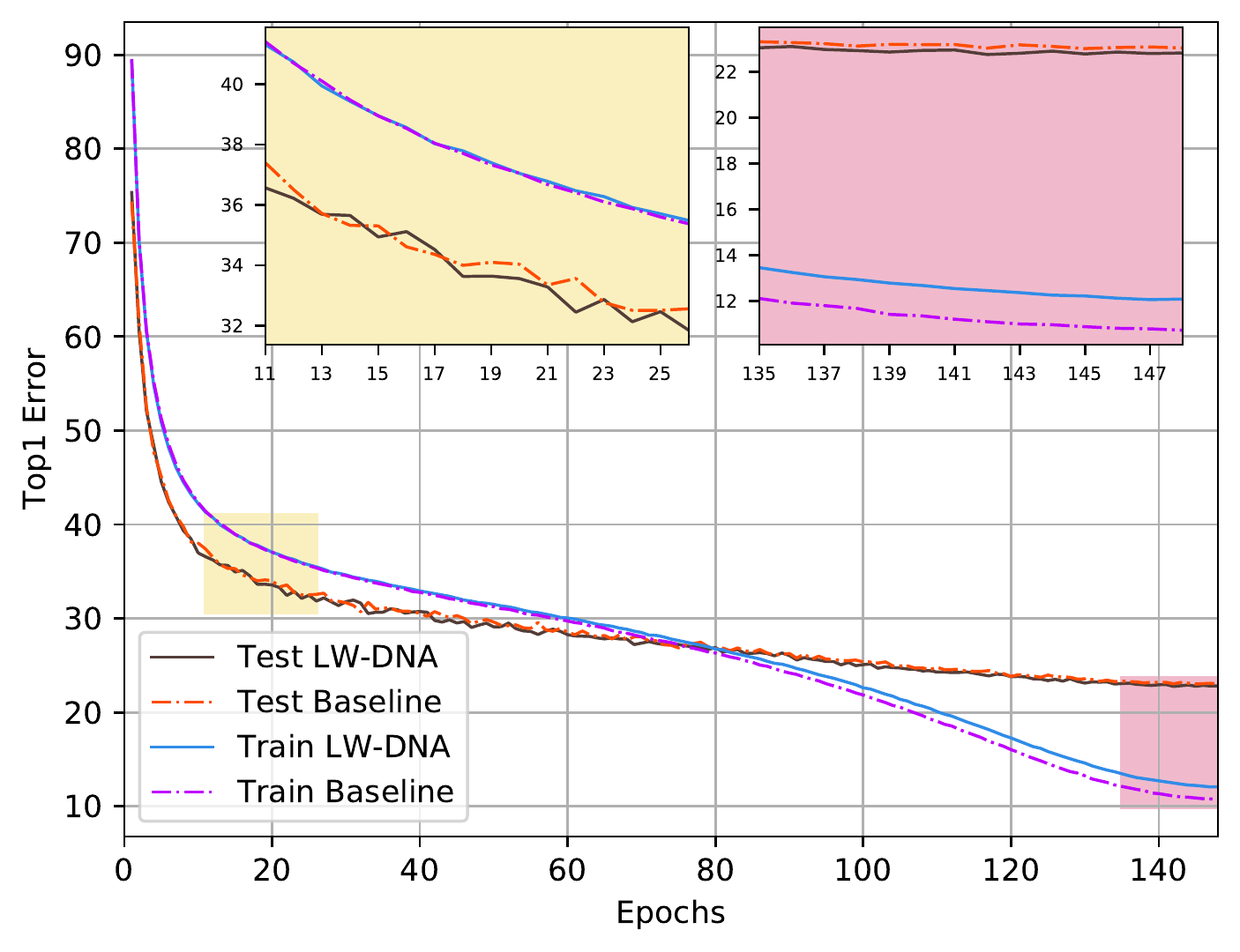}
  \vspace{-0.6cm}
  \subcaption{RegNet-4GF, ImageNet.}
  \label{fig:log_regnet_4gf}
\end{minipage}%
\begin{minipage}[c]{0.33\textwidth}
  \vspace*{\fill}
  \centering
  \includegraphics[width=1\linewidth]{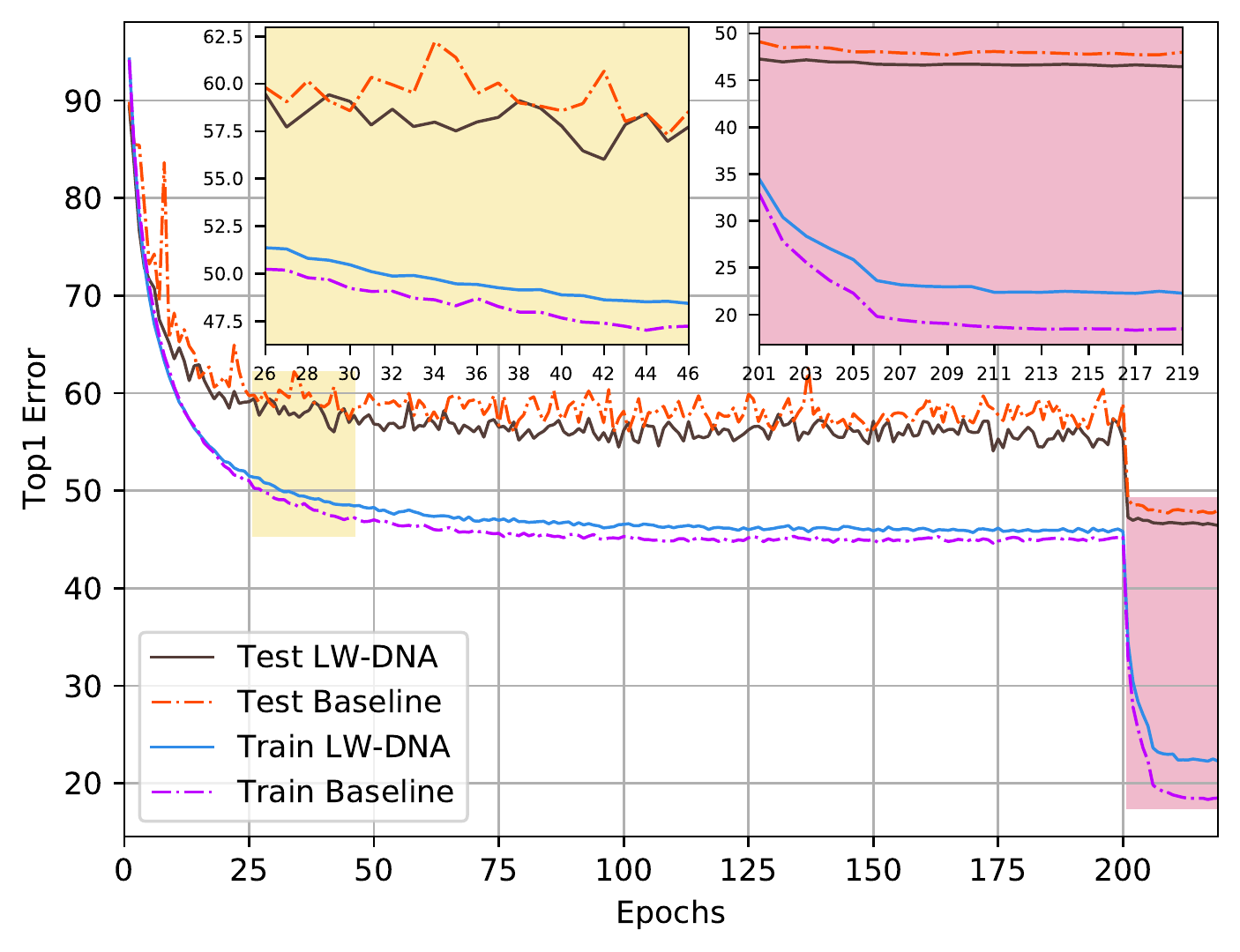}
  \vspace{-0.6cm}
  \subcaption{MobileNetV1, Tiny-ImageNet.}
  \label{fig:log_mobilenetv1}
\end{minipage}%
\vspace{-0.2cm}
\caption{Training and testing log of the LW-DNA models and the baseline models.}
\label{fig:log}
\end{figure*}

\begin{figure*}[!hbt]
\begin{minipage}[c]{0.25\textwidth}
  \vspace*{\fill}
  \centering
  \includegraphics[width=1\linewidth]{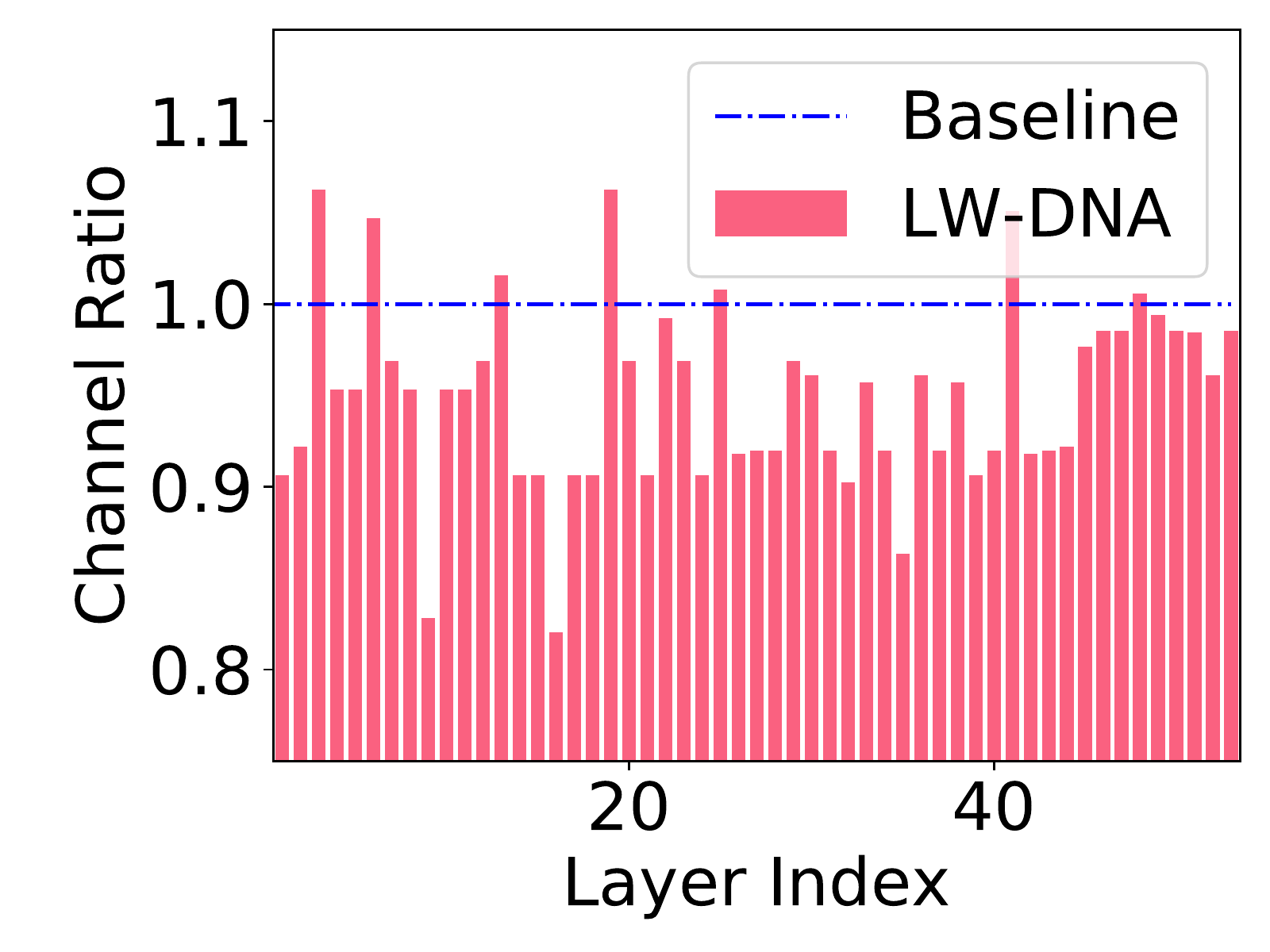}
  \vspace{-0.6cm}
  \subcaption{\small ResNet50, ImageNet.}
  \label{fig:channel_percentage_resnet50}
\end{minipage}%
\begin{minipage}[c]{0.25\textwidth}
  \vspace*{\fill}
  \centering
  \includegraphics[width=1\linewidth]{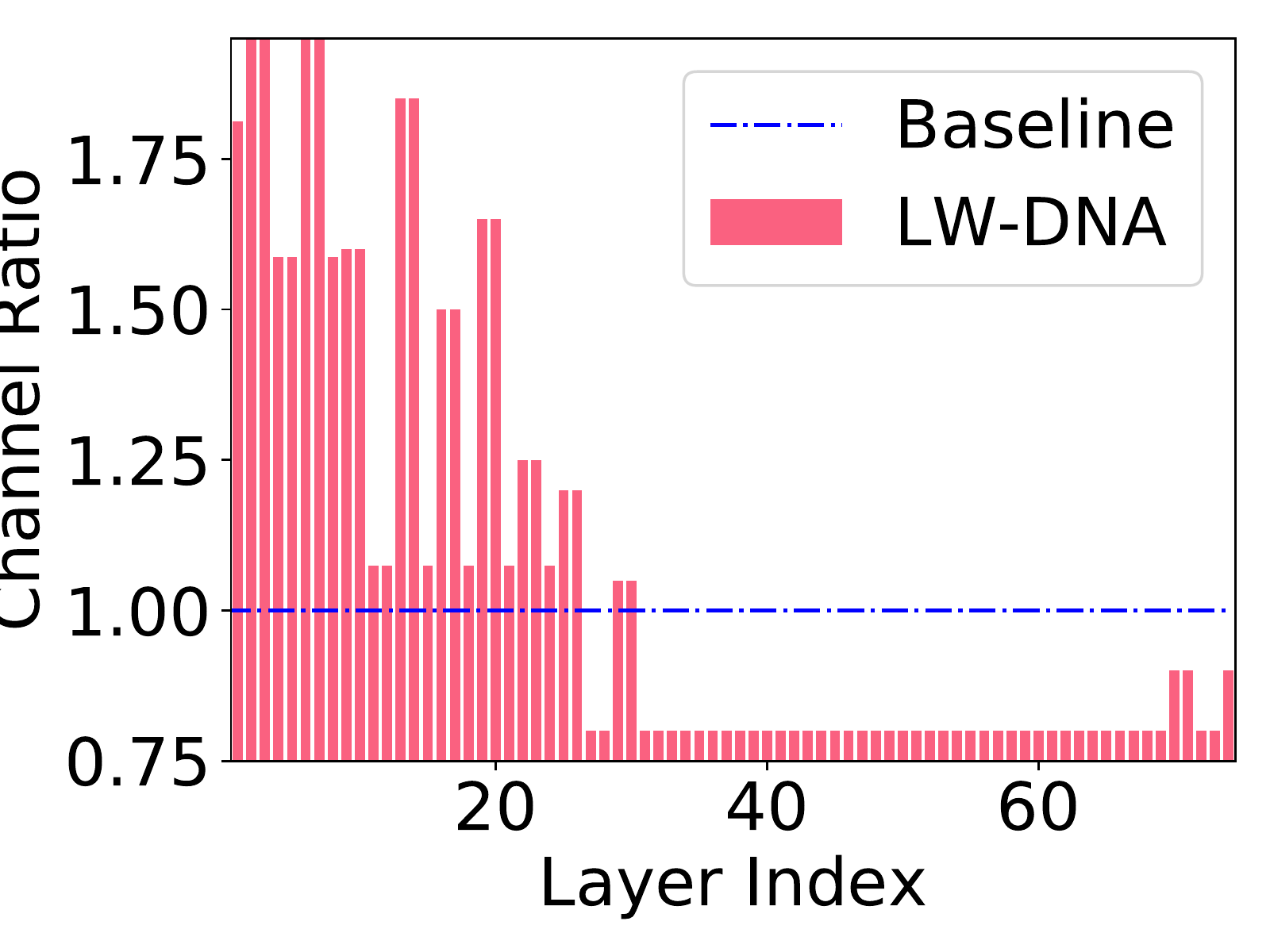}
  \vspace{-0.6cm}
  \subcaption{\small RegNet-4GF, ImageNet.}
  \label{fig:channel_percentage_regnet_4gf}
\end{minipage}%
\begin{minipage}[c]{0.25\textwidth}
  \vspace*{\fill}
  \centering
  \includegraphics[width=1\linewidth]{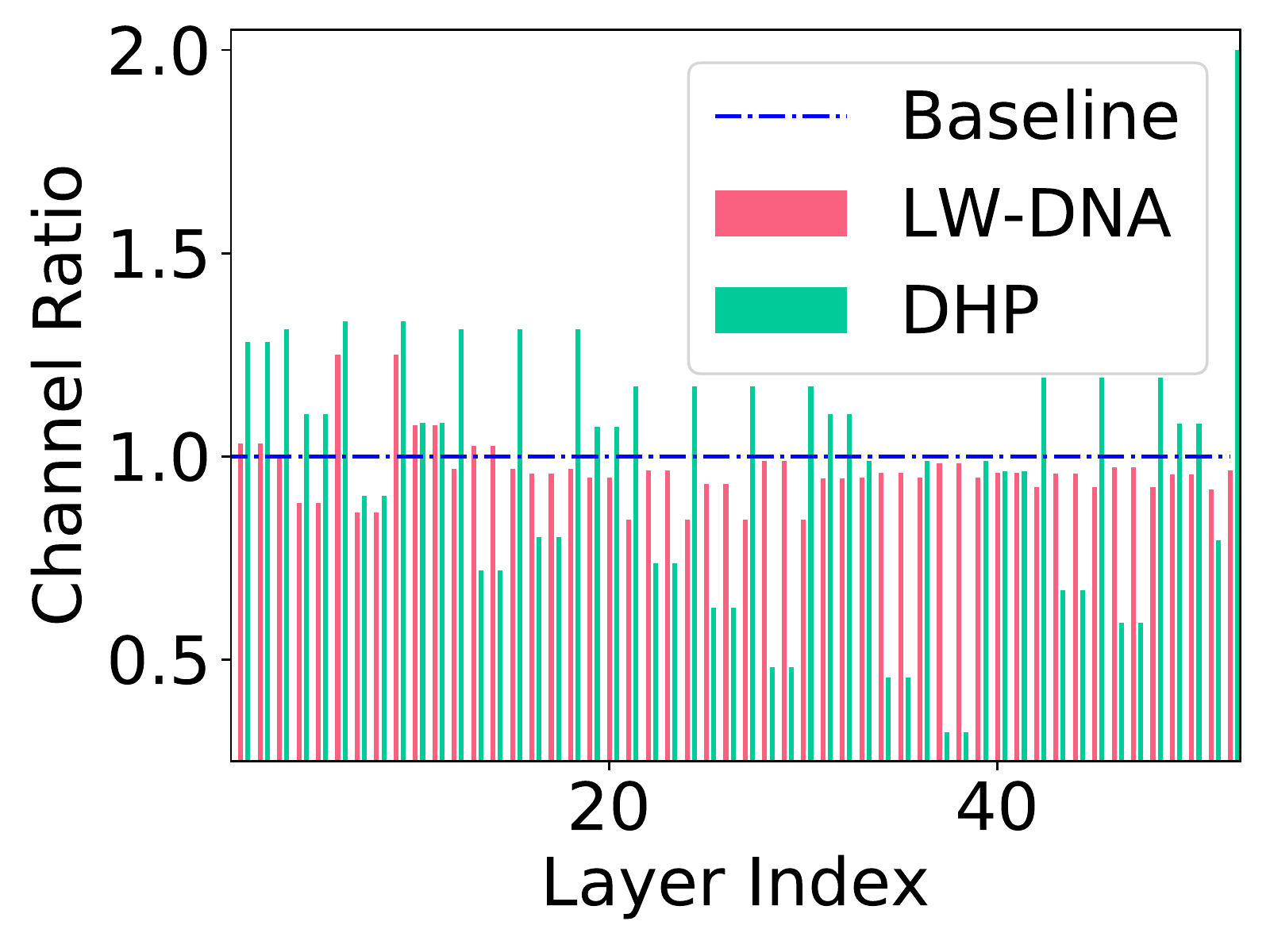}
  \vspace{-0.6cm}
  \subcaption{\small MobileNetV2, Tiny-ImageNet.}
  \label{fig:channel_percentage_mobilenetv2}
\end{minipage}%
\begin{minipage}[c]{0.25\textwidth}
  \vspace*{\fill}
  \centering
  \includegraphics[width=1\linewidth]{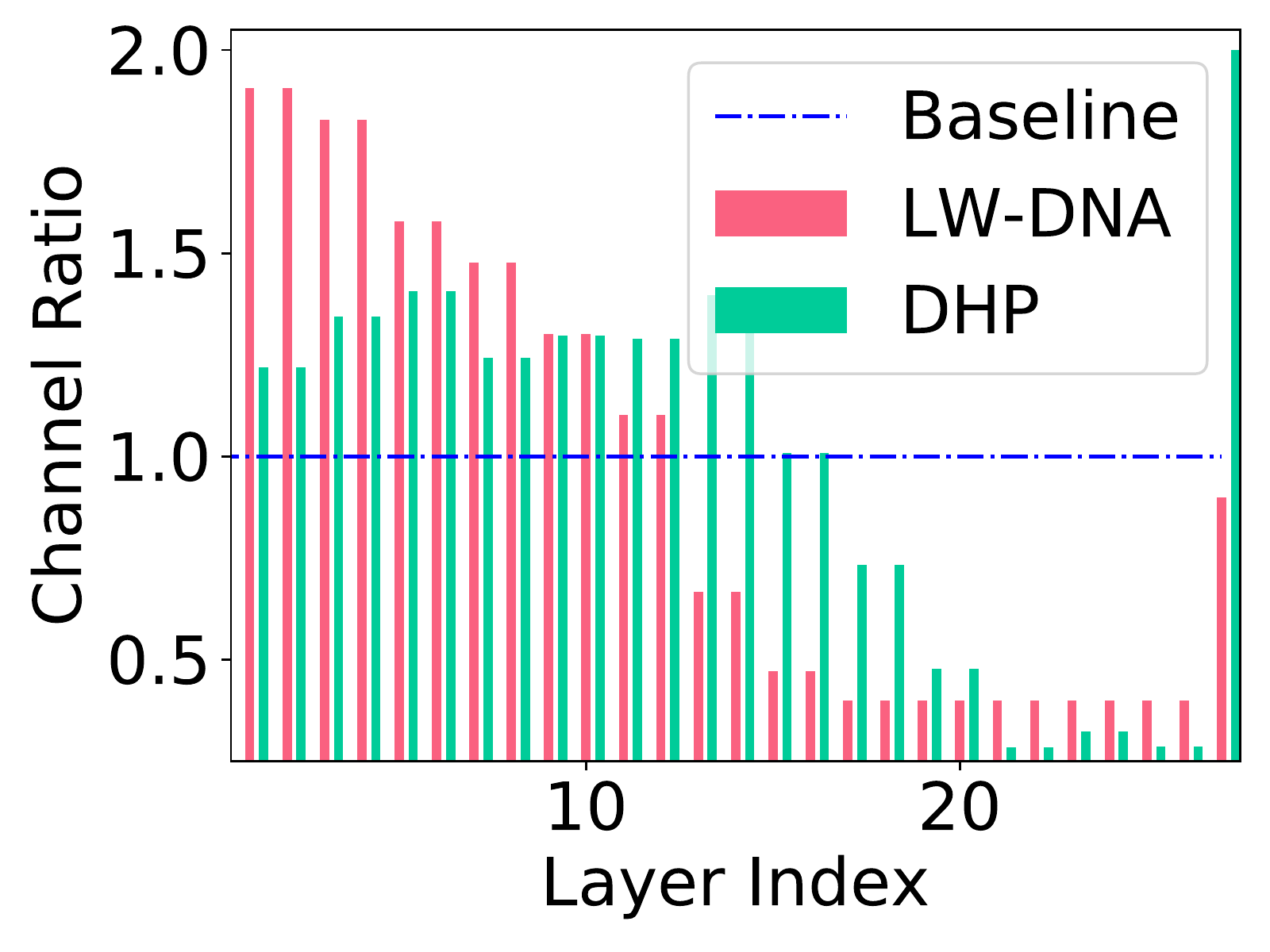}
  \vspace{-0.6cm}
  \subcaption{\small{MobileNetV1, Tiny-ImageNet.}}
  \label{fig:channel_percentage_mobilenetv1}
\end{minipage}%
\vspace{-0.2cm}
\caption{Percentage of remaining output channels of LW-DNA models over the baseline network.}
\label{fig:channel_percentage}
\end{figure*}

The experimental results are shown in this section. We try to identify LW-DNA for various state-of-the-art networks including ResNet~\cite{he2016deep}, RegNet~\cite{radosavovic2020designing}, MobileNets~\cite{howard2017mobilenets,sandler2018mobilenetv2,howard2019searching}, EfficientNet~\cite{tan2019efficientnet}, MnasNet~\cite{tan2019mnasnet}, DenseNet~\cite{huang2017densely}, SRResNet~\cite{ledig2016photo}, EDSR~\cite{lim2017enhanced}, DnCNN~\cite{zhang2017beyond}, and U-Net~\cite{ronneberger2015unet}. The identified LW-DNA model and the baseline network are trained with exactly the same training protocol. The details of the training protocol for different tasks are given in the supplementary. 
Knowledge distillation~\cite{hinton2015distilling} is used for image classification on CIFAR~\cite{krizhevsky2009learning} and Tiny-ImageNet\cite{deng2009imagenet} (Baseline KD, DHP KD~\cite{li2020dhp}, and LW-DNA model). The balancing hyperparameter and temperature are set to 0.4 and 4, respectively. The teacher is the pretrained widened version of the baseline network. Knowledge distillation is not used for experiments on ImageNet because the execution of the teacher network in this case also consumes considerable time and GPU resources.

\textbf{\textit{Image classification.}}
The results of image classification networks are compared in Table~\ref{tbl:results_classification}. A complete version of the results is given in the supplementary. We have several key observations. \textbf{I.} The identified LW-DNA models outperform the original network (denoted as Baseline or Baseline KD when knowledge distillation is used) with lower model complexity in terms of both FLOPs and number of parameters. This is a direct support for the Heterogeneity Hypothesis. \textbf{II.} The accuracy of the baseline network can be improved by knowledge distillation. Yet, the improved baseline still performs worse than LW-DNA. This shows the robustness of LW-DNA, \ie not affected by a specific training technique. \textbf{III.} The improvement of LW-DNA scales up to large-scale datasets, \ie ImageNet. For the ImageNet experiment, we set $\rho = 0.4$ and $\tau = 0.45$ by the ablation study on Tiny-ImageNet shown in the supplementary. This hyper-parameter combination works well across the three investigated networks. The success on ImageNet and the robustness of the hyper-parameters imply the wide existence of LW-DNA models and the ease of finding them. \textbf{IV.} MutualNet is a training scheme when applied to a specific network, which could be combined with our work.

\textbf{\textit{Proximal gradient descent \vs single-shot shrinkage.}}
Besides single-shot shrinkage, there are also other candidate methods to prune networks, \eg proximal gradient descent (PGD).
The choice of single-shot shrinkage is based on the following considerations. 
First, it is extremely computation-efficient. Only one random batch is used to identify the LW-DNA models. This meets the design requirements of introducing no computational cost.
This consistence makes it possible to identify the importance of the architecture of LW-DNA models while controlling the other factors. 
Secondly, by analyzing the closed-form solution to the proximal operator with $\ell_1$ regularization, \ie the soft-thresholding operator, we find that PGD tends to diminish the elements of the latent vectors with the approximately consistent speed. 
As a results, the final magnitude of the elements has some kind of relationship with the initial magnitude.
Therefore, if the initialization of an element is large, it is likely that the final magnitude is still relatively large. 
The distribution of the latent vectors during the PGD optimization is shown in the supplementary. The final distribution is related to the initialization. Thus, it becomes reasonable to shrink the latent vectors at initialization.

\begin{table*}[!t]
    \small
    \begin{center}
        \begin{tabular}{c|c|c|c|c|c|c|c|c}
            \toprule
            \multirow{2}{*}{Network} & \multirow{2}{*}{Method} & \multicolumn{5}{|c|}{PSNR $[dB]$} & \multirow{2}{*}{\shortstack{FLOPs $[G]$ /\\ Ratio (\%)}} & \multirow{2}{*}{\shortstack{Params $[M]$ /\\Ratio (\%)}}  \\ \cline{3-7}
            &  & { Set5~\cite{bevilacqua2012low}} & { Set14~\cite{zeyde2010single}} & { B100~\cite{MartinFTM01}} & { Urban100~\cite{Huang-CVPR-2015}} & { DIV2K~\cite{Agustsson_2017_CVPR_Workshops}} &  & \\ \midrule
            \multirow{2}{*}{SRResNet~\cite{ledig2016photo}} & Baseline  & 32.02  & 28.50  & 27.52  & 25.88  & 28.84 & 32.81 / 100.0 & 1.53 / 100.0 \\
            & LW-DNA & 32.07 & 28.51 & 27.52 & 25.88 & 28.85 & 28.79 / 87.75 & 1.36 / 88.43 \\
            \midrule
            \multirow{2}{*}{EDSR~\cite{lim2017enhanced}}    & Baseline  & 32.10 & 28.55  & 27.55  & 26.02  & 28.93 & 90.37 / 100.0  & 3.70 / 100.0 \\ 
            & LW-DNA & 32.13 &	28.61 &	27.59 &	26.09 &	28.99 & 55.44 / 61.34 &	2.84 / 76.94 \\
            \bottomrule
        \end{tabular}
    \end{center}
    \vspace{-0.4cm}
    \caption{Results on single image super-resolution networks. The upscaling factor is $\times 4$.}
    \label{tbl:results_super_resolution}
    \vspace{-0.4cm}
\end{table*}

\textbf{\textit{The benefits of LW-DNA models}} are analyzed by several observations of the experimental results. \textbf{I.} The percentage of remaining channels is shown in Fig.~\ref{fig:channel_percentage}. Some layers of the LW-DNA networks are strengthened. This might contribute to the improved performance of LW-DNA. \textbf{II.} As shown in Fig.~\ref{fig:log}, towards the end of the training, the LW-DNA models shoot a lower test error with increased training error. The improved generalization on the test set comes with reduced model complexity and lower training accuracy. This phenomenon is consistent with the pioneering unstructured pruning methods~\cite{lecun1990optimal,hassibi1993second} that try to balance model complexity and overfitting. The same phenomenon on both unstructured pruning and structured pruning points to a common underlying factor. \textbf{III.} The accuracy gain of LW-DNA on Tiny-ImageNet is larger than ImageNet. As known, smaller datasets are easier to be overfitted to. \textbf{IV.} On ImageNet, it is easier to identify LW-DNA models for ResNet50 and RegNet than MobileNetV3. Since the larger models ResNet50 and RegNet contain more redundancy, it is easier for them to overfit the dataset. Based on the above observations, we conjecture that the improvement of LW-DNA model might be related to model overfitting.

\begin{figure}[!t]
    \centering
    \includegraphics[width=0.7\linewidth]{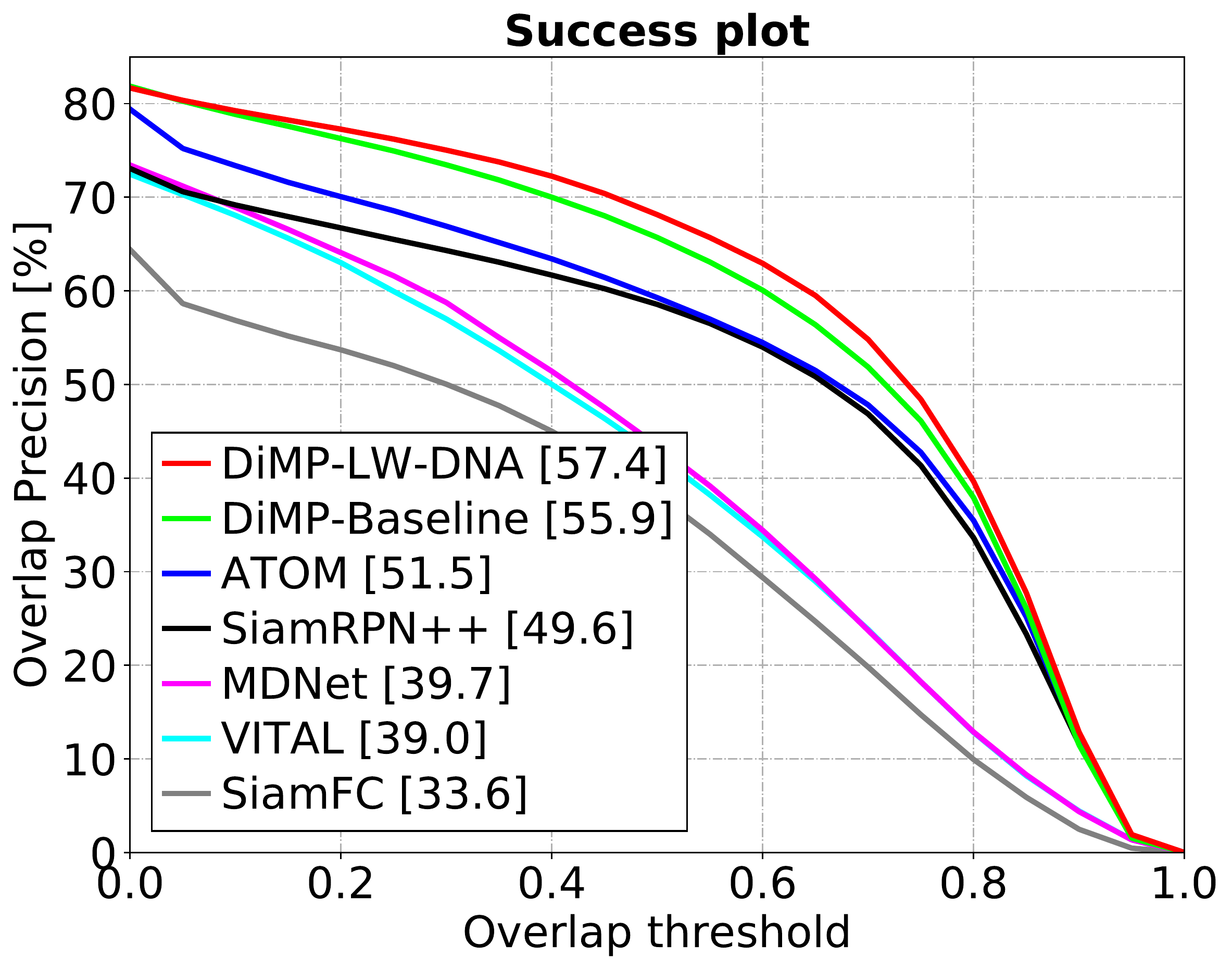}
    \vspace{-0.2cm}
    \caption{Success plot on the LaSOT dataset for visual tracking.}
    \label{fig:results_tracking}
    \vspace{-0.4cm}
\end{figure}

\textbf{\textit{Visual tracking.}} To validate the generalization ability of the identified LW-DNA, we apply the LW-DNA and baseline version of ResNet50 to visual tracking. State-of-the-art tracking workflow DiMP~\cite{bhat2019learning} is used as the test bed. For a fair comparison, the LW-DNA and the baseline are trained with the same protocol. They are first pretrained on ImageNet then finetuned following the DiMP workflow.
In Table~\ref{tbl:results_tracking}, the networks are compared on two datasets, \ie TrackingNet~\cite{muller2018trackingnet} and LaSOT~\cite{fan2019lasot}. On the smaller dataset TrackingNet, LW-DNA version slightly beats the baseline while on the larger dataset LaSOT, LW-DNA outperforms the baseline elegantly. The success plot on LaSOT is shown in Fig.~\ref{fig:results_tracking}. As shown there, DiMP-LW-DNA is consistently better than DiMP-Baseline and other state-of-the-art tracking methods across the range of overlap threshold. In conclusion, the results show that \textbf{the benefits of LW-DNA can be transferred to other vision tasks}.

\textbf{\textit{Image Restoration.}}
Table~\ref{tbl:results_super_resolution} shows the results on super-resolution networks. For EDSR, the LW-DNA models perform better than the baseline but with significant reduction of model complexity. On the large test dataset Urban100 and DIV2K, the LW-DNA model of EDSR leads to nearly 0.1dB PSNR gain over the baseline. For SRResNet, LW-DNA achieves slightly reduction of model complexity without drop of PSNR. More results on image denoising are shown in the supplementary. In conclusion, the results \textbf{validate the existence of LW-DNA models for low-level vision networks}.

\begin{table}[!t]
    \centering
    \small
    \begin{tabular}{c|c|c}
        \toprule
        Metric & DiMP-Baseline & DiMP-LW-DNA \\ \midrule
        \multicolumn{3}{c}{TrackingNet~\cite{muller2018trackingnet}} \\ \midrule
        Precision & 68.06 & 68.27 \\
        Norm. Prec. (\%) & 79.70 & 79.64 \\
        Success (AUC) (\%) & 73.77 & 73.83 \\ \midrule
        \multicolumn{3}{c}{LaSOT~\cite{fan2019lasot}} \\ \midrule
        Precision & 54.97 & 57.30 \\
        Norm. Prec. (\%) & 63.70 & 65.82 \\
        Success (AUC) (\%) & 55.87 & 57.43 \\
        \bottomrule
    \end{tabular}
    \caption{Tracking test results. DiMP-LW-DNA and DiMP-Baseline use the identified LW-DNA and baseline version of ResNet50, respectively.}
    \label{tbl:results_tracking}
    \vspace{-0.4cm}
\end{table}

\section{Conclusion}
In this paper, we state the heterogeneity hypothesis which in essence is the existence of advantageous LW-DNA models for a predefined network architecture. We try to validate the hypothesis by empirical studies. In order to single out the importance of the network architecture, the training protocol is kept the same for the baseline and the LW-DNA models. This is achieved by converting the problem of identifying LW-DNA to a network shrinkage problem and designing an efficient shrinkage algorithm. The experiments on various network architectures and vision tasks demonstrate the benefits of the identified LW-DNA models. By examining the results, we conjecture that the advantage of the LW-DNA model might be related to model overfitting. 

{\footnotesize \noindent \textbf{Acknowledgement:} This work is partially supported by the Major Project for New Generation of AI under Grant No. 2018AAA0100400, the ETH Z\"urich Fund (OK), a Huawei Technologies Oy (Finland) project, an Amazon AWS grant, and an Nvidia grant.}

{\small
\bibliographystyle{ieee_fullname}
\bibliography{ms}

\begin{thebibliography}{10}\itemsep=-1pt

\bibitem{Agustsson_2017_CVPR_Workshops}
Eirikur Agustsson and Radu Timofte.
\newblock {NTIRE} 2017 challenge on single image super-resolution: Dataset and
  study.
\newblock In {\em Proc. CVPRW}, July 2017.

\bibitem{bevilacqua2012low}
Marco Bevilacqua, Aline Roumy, Christine Guillemot, and Marie~Line
  Alberi-Morel.
\newblock Low-complexity single-image super-resolution based on nonnegative
  neighbor embedding.
\newblock In {\em Proc. BMVC}, 2012.

\bibitem{bhat2019learning}
Goutam Bhat, Martin Danelljan, Luc~Van Gool, and Radu Timofte.
\newblock Learning discriminative model prediction for tracking.
\newblock In {\em Proc. ICCV}, pages 6182--6191, 2019.

\bibitem{chang2020principled}
Oscar Chang, Lampros Flokas, and Hod Lipson.
\newblock Principled weight initialization for hypernetworks.
\newblock In {\em Proc. ICLR}, 2020.

\bibitem{chen2019progressive}
Xin Chen, Lingxi Xie, Jun Wu, and Qi Tian.
\newblock Progressive differentiable architecture search: Bridging the depth
  gap between search and evaluation.
\newblock In {\em Proceedings of the IEEE/CVF International Conference on
  Computer Vision}, pages 1294--1303, 2019.

\bibitem{deng2009imagenet}
Jia Deng, Wei Dong, Richard Socher, Li-Jia Li, Kai Li, and Li Fei-Fei.
\newblock Image{N}et: A large-scale hierarchical image database.
\newblock In {\em Proc. CVPR}, pages 248--255. IEEE, 2009.

\bibitem{ding2020lossless}
Xiaohan Ding, Tianxiang Hao, Ji Liu, Jungong Han, Yuchen Guo, and Guiguang
  Ding.
\newblock Lossless cnn channel pruning via gradient resetting and convolutional
  re-parameterization.
\newblock {\em arXiv preprint arXiv:2007.03260}, 2020.

\bibitem{fan2019lasot}
Heng Fan, Liting Lin, Fan Yang, Peng Chu, Ge Deng, Sijia Yu, Hexin Bai, Yong
  Xu, Chunyuan Liao, and Haibin Ling.
\newblock La{SOT}: A high-quality benchmark for large-scale single object
  tracking.
\newblock In {\em Proc. CVPR}, pages 5374--5383, 2019.

\bibitem{frankle2018lottery}
Jonathan Frankle and Michael Carbin.
\newblock The lottery ticket hypothesis: Finding sparse, trainable neural
  networks.
\newblock {\em arXiv preprint arXiv:1803.03635}, 2018.

\bibitem{guo2020single}
Zichao Guo, Xiangyu Zhang, Haoyuan Mu, Wen Heng, Zechun Liu, Yichen Wei, and
  Jian Sun.
\newblock Single path one-shot neural architecture search with uniform
  sampling.
\newblock In {\em European Conference on Computer Vision}, pages 544--560.
  Springer, 2020.

\bibitem{ha2017hypernetworks}
David Ha, Andrew Dai, and Quoc~V Le.
\newblock Hyper{N}etworks.
\newblock In {\em Proc. ICLR}, 2017.

\bibitem{han2015deep}
Song Han, Huizi Mao, and William~J Dally.
\newblock Deep compression: Compressing deep neural networks with pruning,
  trained quantization and {H}uffman coding.
\newblock In {\em Proc. ICLR}, 2015.

\bibitem{hassibi1993second}
Babak Hassibi and David~G Stork.
\newblock Second order derivatives for network pruning: Optimal brain surgeon.
\newblock In {\em Proc. NeurIPS}, pages 164--171, 1993.

\bibitem{he2016deep}
Kaiming He, Xiangyu Zhang, Shaoqing Ren, and Jian Sun.
\newblock Deep residual learning for image recognition.
\newblock In {\em Proc. CVPR}, pages 770--778, 2016.

\bibitem{hinton2015distilling}
Geoffrey Hinton, Oriol Vinyals, and Jeff Dean.
\newblock Distilling the knowledge in a neural network.
\newblock {\em arXiv preprint arXiv:1503.02531}, 2015.

\bibitem{howard2019searching}
Andrew Howard, Mark Sandler, Grace Chu, Liang-Chieh Chen, Bo Chen, Mingxing
  Tan, Weijun Wang, Yukun Zhu, Ruoming Pang, Vijay Vasudevan, et~al.
\newblock Searching for mobilenetv3.
\newblock In {\em Proc. ICCV}, pages 1314--1324, 2019.

\bibitem{howard2017mobilenets}
Andrew~G Howard, Menglong Zhu, Bo Chen, Dmitry Kalenichenko, Weijun Wang,
  Tobias Weyand, Marco Andreetto, and Hartwig Adam.
\newblock Mobile{N}ets: Efficient convolutional neural networks for mobile
  vision applications.
\newblock {\em arXiv preprint arXiv:1704.04861}, 2017.

\bibitem{huang2017densely}
Gao Huang, Zhuang Liu, Laurens van~der Maaten, and Kilian~Q Weinberger.
\newblock Densely connected convolutional networks.
\newblock In {\em Proc. CVPR}, pages 2261--2269, 2017.

\bibitem{Huang-CVPR-2015}
Jia-Bin Huang, Abhishek Singh, and Narendra Ahuja.
\newblock Single image super-resolution from transformed self-exemplars.
\newblock In {\em Proc. CVPR}, pages 5197--5206, 2015.

\bibitem{krizhevsky2009learning}
Alex Krizhevsky and Geoffrey Hinton.
\newblock Learning multiple layers of features from tiny images.
\newblock Technical report, Citeseer, 2009.

\bibitem{lecun1998gradient}
Yann LeCun, L{\'e}on Bottou, Yoshua Bengio, and Patrick Haffner.
\newblock Gradient-based learning applied to document recognition.
\newblock {\em Proceedings of the IEEE}, 86(11):2278--2324, 1998.

\bibitem{lecun1990optimal}
Yann LeCun, John~S Denker, and Sara~A Solla.
\newblock Optimal brain damage.
\newblock In {\em Proc. NeurIPS}, pages 598--605, 1990.

\bibitem{ledig2016photo}
Christian Ledig, Lucas Theis, Ferenc Husz{\'a}r, Jose Caballero, Andrew
  Cunningham, Alejandro Acosta, Andrew Aitken, Alykhan Tejani, Johannes Totz,
  Zehan Wang, et~al.
\newblock Photo-realistic single image super-resolution using a generative
  adversarial network.
\newblock In {\em Proc. CVPR}, pages 105--114, 2017.

\bibitem{lee2019signal}
Namhoon Lee, Thalaiyasingam Ajanthan, Stephen Gould, and Philip~HS Torr.
\newblock A signal propagation perspective for pruning neural networks at
  initialization.
\newblock {\em arXiv preprint arXiv:1906.06307}, 2019.

\bibitem{lee2018snip}
Namhoon Lee, Thalaiyasingam Ajanthan, and Philip~HS Torr.
\newblock {SNIP}: Single-shot network pruning based on connection sensitivity.
\newblock {\em arXiv preprint arXiv:1810.02340}, 2018.

\bibitem{li2017pruning}
Hao Li, Asim Kadav, Igor Durdanovic, Hanan Samet, and Hans~Peter Graf.
\newblock Pruning filters for efficient convnets.
\newblock In {\em Proc. ICLR}, 2017.

\bibitem{li2020group}
Yawei Li, Shuhang Gu, Christoph Mayer, Luc Van~Gool, and Radu Timofte.
\newblock Group sparsity: The hinge between filter pruning and decomposition
  for network compression.
\newblock In {\em Proc. CVPR}, 2020.

\bibitem{li2020dhp}
Yawei Li, Shuhang Gu, Kai Zhang, Luc Van~Gool, and Radu Timofte.
\newblock {DHP}: Differentiable meta pruning via hypernetworks.
\newblock {\em arXiv preprint arXiv:2003.13683}, 2020.

\bibitem{lim2017enhanced}
Bee Lim, Sanghyun Son, Heewon Kim, Seungjun Nah, and Kyoung~Mu Lee.
\newblock Enhanced deep residual networks for single image super-resolution.
\newblock In {\em Proc. CVPRW}, pages 1132--1140, 2017.

\bibitem{liu2018nas}
Chenxi Liu, Barret Zoph, Maxim Neumann, Jonathon Shlens, Wei Hua, Li-Jia Li, Li
  Fei-Fei, Alan Yuille, Jonathan Huang, and Kevin Murphy.
\newblock Progressive neural architecture search.
\newblock In {\em Proc. ECCV}, September 2018.

\bibitem{liu2019darts}
Hanxiao Liu, Karen Simonyan, and Yiming Yang.
\newblock {DARTS}: Differentiable architecture search.
\newblock In {\em Proc. ICLR}, 2019.

\bibitem{liu2019metapruning}
Zechun Liu, Haoyuan Mu, Xiangyu Zhang, Zichao Guo, Xin Yang, Tim Kwang-Ting
  Cheng, and Jian Sun.
\newblock Meta{P}runing: Meta learning for automatic neural network channel
  pruning.
\newblock In {\em Proc. ICCV}, 2019.

\bibitem{ma2018shufflenet}
Ningning Ma, Xiangyu Zhang, Hai-Tao Zheng, and Jian Sun.
\newblock Shuffle{N}et {V}2: Practical guidelines for efficient cnn
  architecture design.
\newblock In {\em Proc ECCV}, pages 116--131, 2018.

\bibitem{malach2020proving}
Eran Malach, Gilad Yehudai, Shai Shalev-Schwartz, and Ohad Shamir.
\newblock Proving the lottery ticket hypothesis: Pruning is all you need.
\newblock In {\em International Conference on Machine Learning}, pages
  6682--6691. PMLR, 2020.

\bibitem{MartinFTM01}
D. Martin, C. Fowlkes, D. Tal, and J. Malik.
\newblock A database of human segmented natural images and its application to
  evaluating segmentation algorithms and measuring ecological statistics.
\newblock In {\em Proc. ICCV}, volume~2, pages 416--423, July 2001.

\bibitem{muller2018trackingnet}
Matthias Muller, Adel Bibi, Silvio Giancola, Salman Alsubaihi, and Bernard
  Ghanem.
\newblock Trackingnet: A large-scale dataset and benchmark for object tracking
  in the wild.
\newblock In {\em Proc. ECCV}, pages 300--317, 2018.

\bibitem{paszke2017automatic}
Adam Paszke, Sam Gross, Soumith Chintala, Gregory Chanan, Edward Yang, Zachary
  DeVito, Zeming Lin, Alban Desmaison, Luca Antiga, and Adam Lerer.
\newblock Automatic differentiation in {P}y{T}orch.
\newblock 2017.

\bibitem{pham2018efficient}
Hieu Pham, Melody Guan, Barret Zoph, Quoc Le, and Jeff Dean.
\newblock Efficient neural architecture search via parameters sharing.
\newblock In {\em Proc. ICML}, pages 4095--4104, 2018.

\bibitem{radosavovic2020designing}
Ilija Radosavovic, Raj~Prateek Kosaraju, Ross Girshick, Kaiming He, and Piotr
  Doll{\'a}r.
\newblock Designing network design spaces.
\newblock {\em arXiv preprint arXiv:2003.13678}, 2020.

\bibitem{ramanujan2020s}
Vivek Ramanujan, Mitchell Wortsman, Aniruddha Kembhavi, Ali Farhadi, and
  Mohammad Rastegari.
\newblock What's hidden in a randomly weighted neural network?
\newblock In {\em Proceedings of the IEEE/CVF Conference on Computer Vision and
  Pattern Recognition}, pages 11893--11902, 2020.

\bibitem{renda2020comparing}
Alex Renda, Jonathan Frankle, and Michael Carbin.
\newblock Comparing rewinding and fine-tuning in neural network pruning.
\newblock {\em arXiv preprint arXiv:2003.02389}, 2020.

\bibitem{ronneberger2015unet}
Olaf Ronneberger, Philipp Fischer, and Thomas Brox.
\newblock U-{N}et: Convolutional networks for biomedical image segmentation.
\newblock In {\em Proc. MICCAI}, pages 234--241. Springer, 2015.

\bibitem{ru2020neural}
Binxin Ru, Pedro Esperanca, and Fabio Carlucci.
\newblock Neural architecture generator optimization.
\newblock {\em arXiv preprint arXiv:2004.01395}, 2020.

\bibitem{sandler2018mobilenetv2}
Mark Sandler, Andrew Howard, Menglong Zhu, Andrey Zhmoginov, and Liang-Chieh
  Chen.
\newblock Mobile{N}et{V}2: Inverted residuals and linear bottlenecks.
\newblock In {\em Proc. CVPR}, pages 4510--4520, 2018.

\bibitem{simonyan2014very}
Karen Simonyan and Andrew Zisserman.
\newblock Very deep convolutional networks for large-scale image recognition.
\newblock {\em arXiv preprint arXiv:1409.1556}, 2014.

\bibitem{tan2019mnasnet}
Mingxing Tan, Bo Chen, Ruoming Pang, Vijay Vasudevan, Mark Sandler, Andrew
  Howard, and Quoc~V Le.
\newblock Mnasnet: Platform-aware neural architecture search for mobile.
\newblock In {\em Proc. CVPR}, pages 2820--2828, 2019.

\bibitem{tan2019efficientnet}
Mingxing Tan and Quoc~V Le.
\newblock Efficientnet: Rethinking model scaling for convolutional neural
  networks.
\newblock {\em arXiv preprint arXiv:1905.11946}, 2019.

\bibitem{xu2019pc}
Yuhui Xu, Lingxi Xie, Xiaopeng Zhang, Xin Chen, Guo-Jun Qi, Qi Tian, and
  Hongkai Xiong.
\newblock {PC-DARTS}: Partial channel connections for memory-efficient
  architecture search.
\newblock {\em arXiv preprint arXiv:1907.05737}, 2019.

\bibitem{yang2020mutualnet}
Taojiannan Yang, Sijie Zhu, Chen Chen, Shen Yan, Mi Zhang, and Andrew Willis.
\newblock Mutual{N}et: Adaptive convnet via mutual learning from network width
  and resolution.
\newblock 2020.

\bibitem{yang2018netadapt}
Tien-Ju Yang, Andrew Howard, Bo Chen, Xiao Zhang, Alec Go, Mark Sandler,
  Vivienne Sze, and Hartwig Adam.
\newblock Net{A}dapt: Platform-aware neural network adaptation for mobile
  applications.
\newblock In {\em Proc. ECCV}, pages 285--300, 2018.

\bibitem{yu2019autoslim}
Jiahui Yu and Thomas Huang.
\newblock Autoslim: Towards one-shot architecture search for channel numbers.
\newblock {\em arXiv preprint arXiv:1903.11728}, 2019.

\bibitem{zeyde2010single}
Roman Zeyde, Michael Elad, and Matan Protter.
\newblock On single image scale-up using sparse-representations.
\newblock In {\em International Conference on Curves and Surfaces}, pages
  711--730. Springer, 2010.

\bibitem{zhang2017beyond}
Kai Zhang, Wangmeng Zuo, Yunjin Chen, Deyu Meng, and Lei Zhang.
\newblock Beyond a {G}aussian denoiser: residual learning of deep {CNN} for
  image denoising.
\newblock {\em IEEE TIP}, 26(7):3142--3155, 2017.

\bibitem{zhou2019deconstructing}
Hattie Zhou, Janice Lan, Rosanne Liu, and Jason Yosinski.
\newblock Deconstructing lottery tickets: Zeros, signs, and the supermask.
\newblock In {\em NeurIPS}, pages 3592--3602, 2019.

\bibitem{zoph2017nas}
Barret Zoph and Quoc~V Le.
\newblock Neural architecture search with reinforcement learning.
\newblock In {\em Proc. ICLR}, 2017.

\end{thebibliography}
}

\appendix
\clearpage


In this supplementary material, we first provide more details about the network shrinkage algorithm in Sec.~\ref{sec:more_details}. Then we describe the training protocols used for training networks on different tasks including image classification, visual tracking, and image restoration in Sec.~\ref{sec:training_protocol}. For a better understanding of the hypernetworks, the demo code is provided in Sec.~\ref{sec:demo_code}. Finally, more experimental results are given in Sec.~\ref{sec:more_experimental_results}.

\section{More Details}
\label{sec:more_details}

\subsection{Handling different layers} 
\label{subsec:handling_different_layers}

In addition to the normal convolutional layers, the hypernetwork can be applied to other layers including depth-wise convolution and batch normalization. 
Special consideration needs to be made for depth-wise convolution. 
For depth-wise convolution, the dimensionality of the weight parameter along the input channel is 1. 
Thus, the latent vector controlling the input channel has only one element and shrinkage of this latent vector is avoided during the network shrinkage phase. 
For the normalization layers like BatchNorm, they are forced to have the same number of output channels as their preceding convolutional layers. The linear layers are pruned along with the preceding convolutional layer such that the same number of channels are removed. 

\subsection{Pruning according to gradients} 
\label{subsec:Pruning_according_gradients}

The proximal gradient descent (PGD) algorithm in DHP~\cite{li2020dhp} prunes the latent vectors according to the magnitude of their elements.
Yet, the problem of PGD algorithm is that it might result in unbalanced pruning. 
For example, in Table~\ref{tbl:results_classification}, for the DHP results on MobileNetV2, the number of parameters increases despite the reduction of FLOPs. 
This is because larger percentage of channels in the lower layers are pruned, which accounts for more FLOPs but less parameters compared with those in the higher layers. 
On the contrary, at initialization of the hypernetwork, the range of the gradients of the latent vectors are relatively balanced across the layers. Thus, gradient magnitude of the latent vectors are chosen as the pruning criterion. 

\section{Training Protocol}
\label{sec:training_protocol}

The code is implemented in PyTorch~\cite{paszke2017automatic}.
For ImageNet experiments, the networks are trained with 4 Nvidia V100 GPUs. For the other experiments, the training is conducted on Nvidia TITAN Xp GPUs. The training protocols for different tasks are explained as follows.

\subsection{Image classification}

\textbf{ImageNet}

The ImageNet2012 dataset has 1000 classes. The training set contains 1.2 million images while the test set contains 50,000 image with 50 image for every class. Standard image normalization and data augmentation method are used. The training continues for 150 epochs. The initial learning rate is 0.05. Cosine learning rate decay is used. The weight decay factor is set to $4e^{-5}$. SGD optimizer is used during the training. The batch size is 256.

\textbf{Tiny-ImageNet}

Tiny-Imagenet is a simplified version of ImageNet2012. It has 200 classes. Each class has 500 training images and 50 validation images. And the resolution of the images is $64 \times 64$. The images are normalized with channel-wise mean and standard deviation. Horizontal flip is used to augment the dataset. The networks are trained for 220 epochs with SGD and an initial learning rate of 0.1. The learning rate is decayed by a factor of 10 at Epoch 200, Epoch 205, Epoch 210, and Epoch 215. The momentum of SGD is 0.9. Weight decay factor is set to 0.0001. The batch size is 64.

\begin{table*}[!hbt]
    \footnotesize
    \begin{center}
        \begin{tabular}{c|c|c|c|c|c}
            \toprule 
             
             Dataset & Network  & Method & Top-1 Error (\%) & FLOPs [G] / Ratio (\%) & Params [M] / Ratio (\%) \\ \midrule
            
            \multirow{9}{*}{ImageNet~\cite{deng2009imagenet}} & \multirow{5}{*}{ResNet50~\cite{he2016deep}}
            &Baseline	&23.28	&4.1177	/ 100.0	&25.557	/ 100.0	\\
            &&MutualNet~\cite{yang2020mutualnet} &21.40 & 4.1177 / 100.0 &25.557 / 100.0\\
            &&LW-DNA	&23.00	&3.7307	/ 90.60	&23.741	/ 92.90	\\ 
            &&MetaPruning~\cite{liu2019metapruning} & 23.80 & 3.0000 / 72.86 & -- \\
            &&AutoSlim~\cite{yu2019autoslim}  &24.00 & 3.0000 / 72.86 & 23.100 / 90.39 \\ \cline{2-6}
            &\multirow{2}{*}{\shortstack{RegNet~\cite{radosavovic2020designing} \\ X-4.0GF}} 
            &Baseline	&23.05	&4.0005	/ 100.0	&22.118	/ 100.0	\\
            &&LW-DNA	&22.74	&3.8199	/ 95.49	&15.285	/ 69.10	\\ \cline{2-6}
            &\multirow{2}{*}{\shortstack{MobileNetV3 small~\cite{howard2019searching}}} 
            &Baseline	&34.91	&0.0612 / 100.0	&3.108	/ 100.0	\\
            &&LW-DNA	&34.84	&0.0605	/ 98.86 &3.049	/ 98.11	\\ \midrule
            
            \multirow{16}{*}{Tiny-ImageNet} & \multirow{4}{*}{MobileNetV1~\cite{howard2017mobilenets}} 
            &Baseline	&51.87	&0.0478	/ 100.0	&3.412	/ 100.0	\\
            &&Baseline KD	&48.00	&0.0478	/ 100.0	&3.412	/ 100.0	\\
            &&DHP KD	&46.70	&0.0474	/ 99.16	&2.267	/ 66.43	\\
            &&LW-DNA	&46.44	&0.0460	/ 96.23	&1.265	/ 37.08	\\ \cline{2-6}
        	&\multirow{4}{*}{MobileNetV2~\cite{sandler2018mobilenetv2}} 	
            &Baseline	&44.38	&0.0930	/ 100.0	&2.480	/ 100.0	\\
            &&Baseline KD	&41.25	&0.0930	/ 100.0	&2.480	/ 100.0	\\
            &&DHP KD	&41.06	&0.0896	/ 96.34	&2.662	/ 107.34	\\
            &&LW-DNA	&40.74	&0.0872	/ 93.76	&2.230	/ 89.90	\\ \cline{2-6}
            &\multirow{4}{*}{\shortstack{MobileNetV3~\cite{howard2019searching} \\ large}} 				
            &Baseline	&45.53	&0.0860	/ 100.0	&4.121	/ 100.0	\\
            &&Baseline KD	&38.21	&0.0860	/ 100.0	&4.121	/ 100.0	\\
            &&DHP KD	&38.14	&0.0856	/ 99.53	&3.561	/ 86.42	\\
            &&LW-DNA	&37.45	&0.0797	/ 92.67	&3.561	/ 86.43	\\ \cline{2-6}
            &\multirow{4}{*}{\shortstack{MobileNetV3~\cite{howard2019searching} \\ small}} 					
            &Baseline	&47.55	&0.0207	/ 100.0	&2.083	/ 100.0	\\
            &&Baseline KD	&41.52	&0.0207	/ 100.0	&2.083	/ 100.0	\\
            &&DHP KD	&41.46	&0.0192	/ 92.75	&1.078	/ 51.76	\\
            &&LW-DNA	&41.35	&0.0178	/ 85.99	&1.799	/ 86.36	\\ \cline{2-6}
            
            &\multirow{4}{*}{MnasNet~\cite{tan2019mnasnet}} 					
            &Baseline	&51.79	&0.0271	/ 100.0	&3.359	/ 100.0	\\
            &&Baseline KD	&48.17	&0.0271	/ 100.0	&3.359	/ 100.0	\\
            &&DHP KD	&48.10	&0.0264	/ 97.42	&2.512	/ 74.79	\\
            &&LW-DNA	&46.85	&0.0250	/ 92.25	&1.258	/ 37.45	\\ \midrule
            
            \multirow{18}{*}{CIFAR100} & \multirow{3}{*}{\shortstack{RegNet~\cite{radosavovic2020designing} \\ Y-200MF}}
            &Baseline	&21.94	&0.2259	/ 100.0	&2.831	/ 100.0	\\
            &&Baseline KD	&19.87	&0.2259	/ 100.0	&2.831	/ 100.0	\\
            &&LW-DNA	&19.87	&0.2095	/ 92.74	&1.524	/ 53.85	\\ \cline{2-6}
            &\multirow{3}{*}{\shortstack{RegNet~\cite{radosavovic2020designing} \\ Y-400MF}}
            &Baseline	&21.65	&0.4585	/ 100.0	&3.947	/ 100.0	\\
            &&Baseline KD	&18.71	&0.4585	/ 100.0	&3.947	/ 100.0	\\
            &&LW-DNA	&18.65	&0.4468	/ 97.45	&2.466	/ 62.48	\\ \cline{2-6}
            &\multirow{3}{*}{\shortstack{RegNet~\cite{radosavovic2020designing} \\ X-200MF}}
            &Baseline	&23.62	&0.2255	/ 100.0	&2.353	/ 100.0	\\
            &&Baseline KD	&21.38	&0.2255	/ 100.0	&2.353	/ 100.0	\\
            &&LW-DNA	&21.19	&0.2075	/ 92.02	&1.239	/ 52.68	\\ \cline{2-6}
            &\multirow{3}{*}{\shortstack{RegNet~\cite{radosavovic2020designing} \\ X-400MF}}
            &Baseline	&21.75	&0.4698	/ 100.0	&4.810	/ 100.0	\\
            &&Baseline KD	&19.06	&0.4698	/ 100.0	&4.810	/ 100.0	\\
            &&LW-DNA	&18.81	&0.4610	/ 98.13	&4.404	/ 91.56	\\ \cline{2-6}
            &\multirow{3}{*}{EfficientNet~\cite{tan2019efficientnet}}					
            &Baseline	&20.74	&0.4161	/ 100.0	&4.136	/ 100.0	\\
            &&Baseline KD	&19.73	&0.4161	/ 100.0	&4.136	/ 100.0	\\
            &&LW-DNA	&19.54	&0.3850	/ 92.53	&2.121	/ 51.28	\\ \cline{2-6}
            &\multirow{3}{*}{DenseNet40~\cite{huang2017densely}}					
            &Baseline	&26.00	&0.2901	/ 100.0	&1.100	/ 100.0	\\
            &&Baseline KD	&22.84	&0.2901	/ 100.0	&1.100	/ 100.0	\\
            &&LW-DNA	&22.46	&0.2638	/ 90.93	&1.016	/ 92.35	\\ \midrule
            
            \multirow{6}{*}{CIFAR10} & \multirow{3}{*}{DenseNet40~\cite{huang2017densely}}
            &Baseline	&5.50	&0.2901	/ 100.0	&1.059	/ 100.0	\\
            &&Baseline KD	&4.88	&0.2901	/ 100.0	&1.059	/ 100.0	\\
            &&LW-DNA	&4.87	&0.2632	/ 90.73	&0.963	/ 90.87	\\ \cline{2-6}
            &\multirow{3}{*}{ResNet56~\cite{he2016deep}}				
            &Baseline	&5.74	&0.1274	/ 100.0	&0.856	/ 100.0	\\
            &&Baseline KD	&5.73	&0.1274	/ 100.0	&0.856	/ 100.0	\\
            &&LW-DNA	&5.49	&0.1262	/ 99.06	&0.536	/ 62.62	\\ \midrule 

        \end{tabular}
    \end{center}
    \caption{Image classification results. Baseline and Baseline KD denote the original network trained without and with knowledge distillation respectively. DHP-KD is the DHP version trained with knowledge distillation.}
    \label{tbl:results_classification}
\end{table*}

\textbf{CIFAR}

CIFAR~\cite{krizhevsky2009learning} dataset contains two datasets \ie CIFAR10 and CIFAR100. CIFAR10 contains 10 different classes. The training subset and testing subset of the the dataset contain 50,000 and 10,000 images with resolution $32 \times 32$, respectively. CIFAR100 is the same as CIFAR10 except that it has 100 classes.  All of the images are normalized using channel-wise mean and standard deviation of the the training set~\cite{he2016deep,huang2017densely}. Standard data augmentation is also applied. Both of the baseline and the LW-DNA networks are trained for 300 epochs with SGD optimizer and an initial learning rate of 0.1. The learning rate is decayed by 10 after 50\% and 75\% of the epochs. The momentum of SGD is 0.9. Weight decay factor is set to 0.0001. The batch size is 64.

\subsection{Visual tracking}

For visual tracking, we follow the training protocol in \cite{bhat2019learning}. To compare the baseline network and the LW-DNA models, the backbone network is initialized with the weights of ResNet50 and LW-DNA trained for this paper, respectively. Then the same training and testing protocol is applied. The results are denoted by DiMP-Baseline and DiMP-LW-DNA respectively.

\subsection{Image restoration}

\textbf{Super-resolution}

DIV2K dataset is used to train image super-resolution networks. The dataset contains 800 training images, 100 validation images, and 100 test images. 
The full resolution images are cropped into $480 \times 480$ subimages with overlap 240. There are 32208 subimages in total.
For EDSR, the size of the extracted low-resolution input patch is $48 \times 48$ while for SRResNet the size is $24 \times 24$. The batch size is 16. Adam optimizer is used for the training. Default hyper-parameters are used for Adam optimizer. The weight decay factor is 0.0001. The networks are trained for 300 epochs. The learning rate starts from 0.0001 and decays by 10 after 200 epochs.

A simplified version of EDSR is used in order to speed up the training of EDSR. The original EDSR network contains 32 residual blocks and each convolutional layer has 256 channels. The simplified version has 8 residual blocks and with 128 channels for each convolutional layers.

\textbf{Denoising}

For image denoising, the images in DIV2K dataset are converted to gray images. For DnCNN the patch size of the input image is $64 \times 64$ and the batch size is 64. For UNet, the patch size and the batch size are $128 \times 128$ and 16, respectively. Gaussian noise is added to degrade the input patches on the fly with noise level $\sigma = 70$. Adam optimizer is used to train the network. The weight decay factor is 0.0001. The networks are trained for 60 epochs and each epoch contains 10,000 iterations. So in total, the training continues for 600k iterations. The learning rate starts with 0.0001 and decays by 10 at Epoch 40.

\section{Demo code of hypernetworks}
\label{sec:demo_code}

\begin{lstlisting}[language=Python, label={lst:hypernetwork}, caption=Demo code of the utilized hypernetworks.]
  import torch
  z_o = torch.randn(n)
  z_i = torch.randn(c)
  w_1 = torch.randn(n, c, m)
  w_2 = torch.randn(n, c, w*h, m)
  o = torch.matmul(z_o.unsqueeze(-1), 
      z_i.unsqueeze(0))
  o = o.unsqueeze(-1) * w_1
  o = torch.matmul(w_2, o.unsqueeze(-1))
\end{lstlisting}

For a better understanding, the demo code of the utilized hypernetworks is shown in the code Listing~\ref{lst:hypernetwork}. The main part of code only contains 3 lines.

\section{More Experimental Results}
\label{sec:more_experimental_results}
\textbf{Full list of image classification results}

Due to the lack of space, only a part of the results on image classification is shown in the main paper. The full list of image classification results is shown in Table~\ref{tbl:results_classification}. Fig.~\ref{fig:log} shows more results on the training and testing log of different models. Fig.~\ref{fig:channel_percentage} shows the percentage of remaining channels of more LW-DNA models.

\textbf{Denoising}

Image denoising results are shown in Table~\ref{tbl:results_denoising}. The identified LW-DNA models perform no worse than the baseline network with reduced number of parameters and FLOPs.
\begin{table*}
    \small
    \begin{center}
        \begin{tabular}{c|c|c|c|c|c}
            \toprule
            \multirow{2}{*}{Network} & \multirow{2}{*}{Method} & \multicolumn{2}{|c|}{{PSNR $[dB]$}} & \multirow{2}{*}{\shortstack[c]{FLOPs $[G]$ /\\Ratio (\%)}} & \multirow{2}{*}{\shortstack[c]{Params $[M]$ /\\Ratio (\%)}}  \\ \cline{3-4} 
            &  & BSD68 & DIV2K && \\ \midrule
            \multirow{2}{*}{DnCNN~\cite{zhang2017beyond}}   & Baseline  & 24.9  & 26.7  & 9.10 / 100.0  & 0.56 / 100.0 \\ 
            & LW-DNA & 24.9 & 26.7 &5.43 / 59.64 & 0.33 / 59.47 \\
            \midrule
            \multirow{2}{*}{U-Net~\cite{ronneberger2015unet}}& Baseline  & 25.2  & 27.2  & 3.41 / 100.0  & 7.76 / 100.0   \\ 
            & LW-DNA &25.2 & 27.2 & 3.26 / 95.60 & 5.86 / 75.57\\
            \bottomrule
        \end{tabular}
    \end{center}
    \caption{Compression results on image denoising networks. The noise level $\sigma$ is 70.}
    \label{tbl:results_denoising}
\end{table*}

\textbf{Ablation study on Tiny-ImageNet}

An ablation study of the hyper-parameters $\rho$ and $\tau$ is shown in Table~\ref{tbl:ablation_study}. The experiments are conducted for MobileNetV1 on Tiny-ImageNet. The FLOPs budget is fixed for the experiments. Two conclusions can be drawn from the result. \textbf{I.} By increasing the hyper-parameters $\rho$ and $\tau$, the model complexity is also increased. And the accuracy of the network is also improved. \textbf{II.} All of the results in Table~\ref{tbl:ablation_study} are better than Baseline KD in Table~\ref{tbl:results_classification}, which shows the robustness of $\rho$ and $\tau$. Based on the experience on Tiny-ImageNet, we set $\rho = 0.4$ and $\tau = 0.45$ for ImageNet experiments. Quite surprising, this combination works well across the three investigated networks (ResNet50, RegNet, and MobileNetV3).

\begin{table*}
    \small
    \begin{center}
        \begin{tabular}{c|c|c|c|c}
        \toprule
        $\rho$ &  $\tau$ &  Top-1 &  FLOPs {[}G{]} &  Params {[}M{]} \\ \hline
        0.1 & 0.4 & 47.02 & 0.046 & 0.948 \\
        0.1 & 0.45 & 46.66 & 0.046 & 0.986 \\
        0.2 & 0.4 & 46.94 & 0.0459 & 1.210 \\
        0.2 & 0.45 & 46.44 & 0.046 & 1.265 \\
        \bottomrule
        \end{tabular}
    \end{center}
    \caption{Ablation study of the hyper-parameters $\rho$ and $\tau$ on MobileNetV1.}
    \label{tbl:ablation_study}
\end{table*}

\begin{figure*}[!hbt]
\begin{minipage}[c]{0.33\textwidth}
  \centering
  \includegraphics[width=1\linewidth]{images/MobileNetV1_Top1_log.pdf}
  \subcaption{MobileNetV2.}
  \label{fig:log_mobilenetv1}
\end{minipage}%
\begin{minipage}[c]{0.33\textwidth}
  \centering
  \includegraphics[width=1\linewidth]{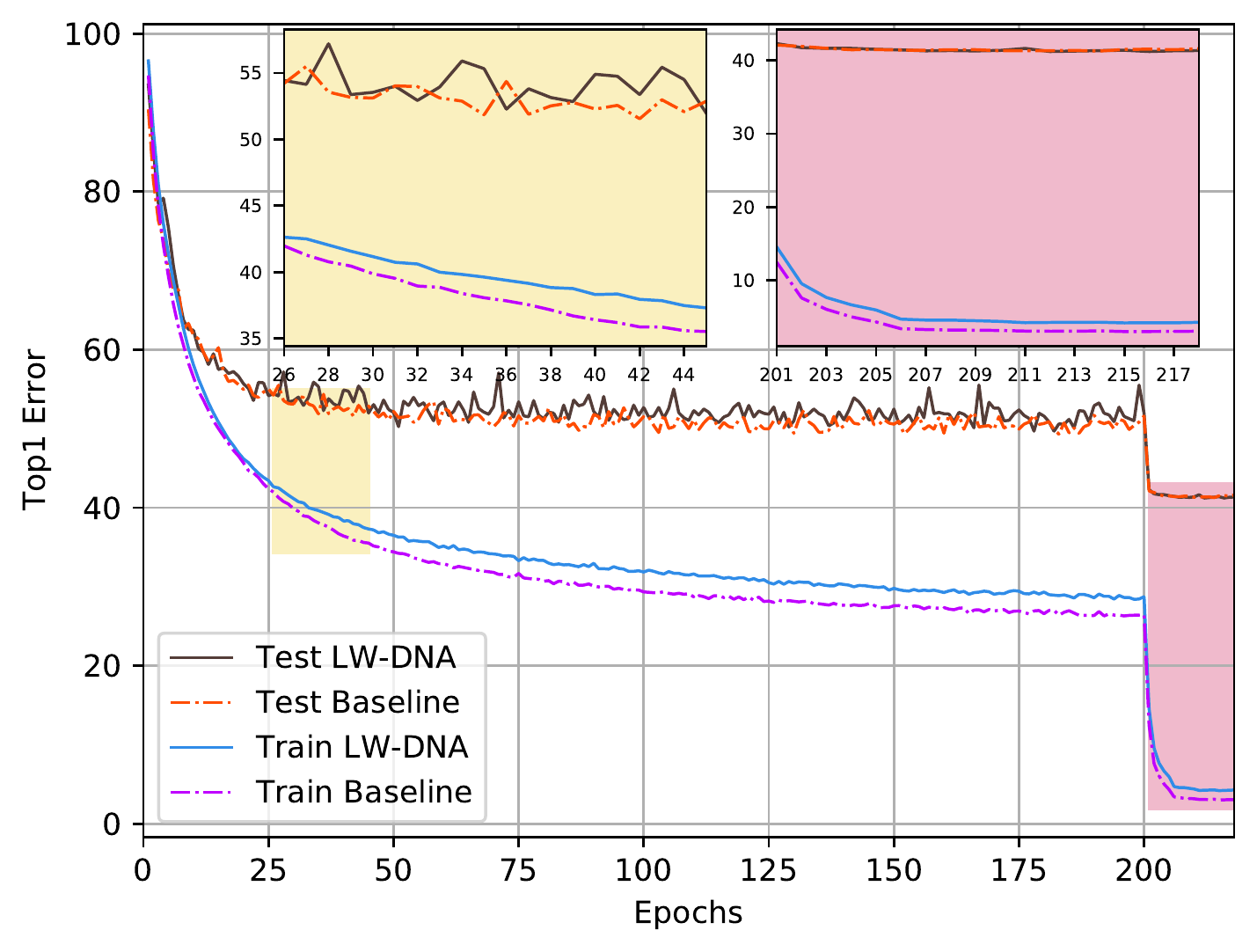}
  \subcaption{MobileNetV3 small.}
  \label{fig:log_mobilenetv3}
\end{minipage}%
\begin{minipage}[c]{0.33\textwidth}
  \centering
  \includegraphics[width=1\linewidth]{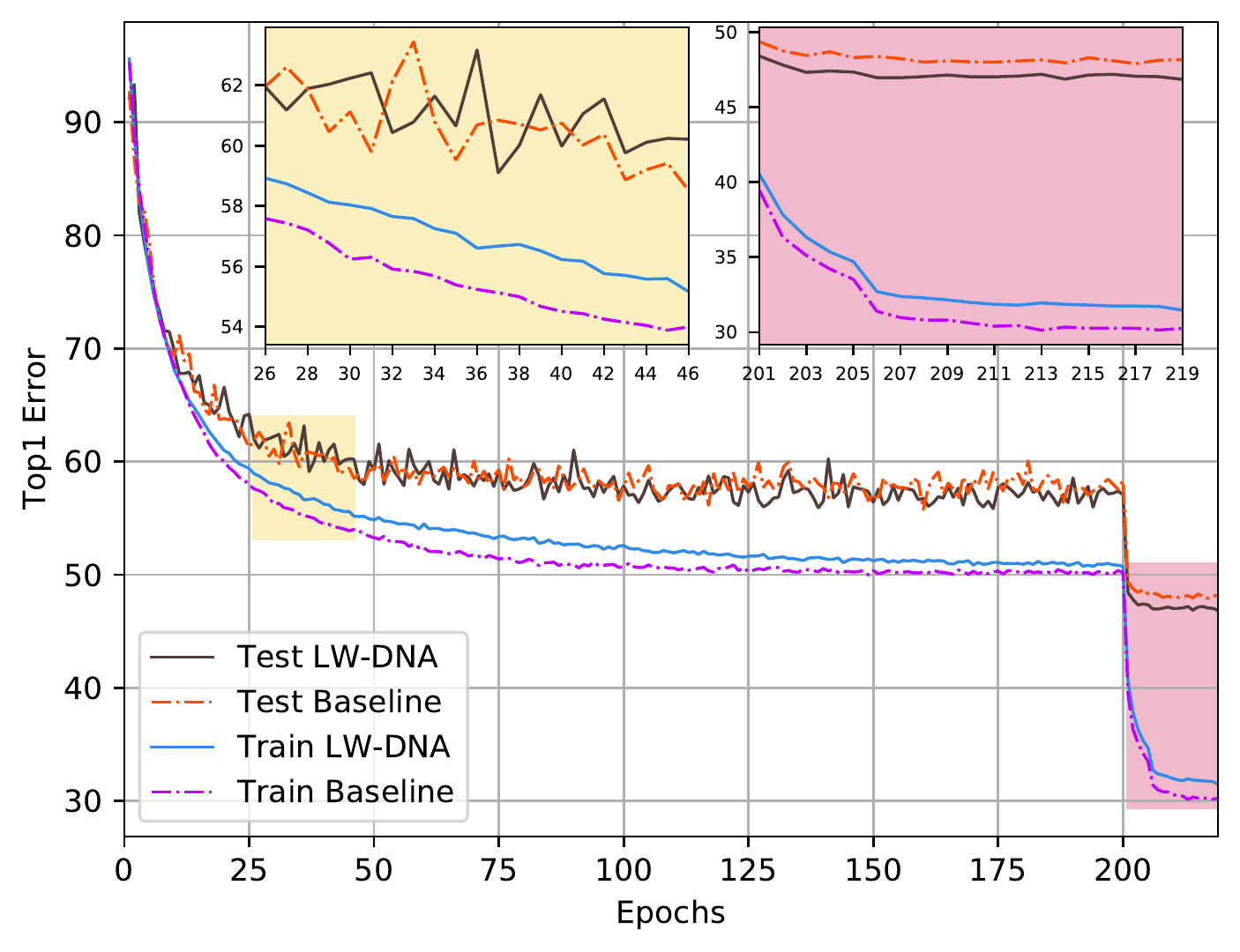}
  \subcaption{MNASNet.}
  \label{fig:log_mnasnet}
\end{minipage}%
\caption{Training and testing log of the LW-DNA models and the baseline models.}
\label{fig:log}
\end{figure*}

\begin{figure*}
\begin{minipage}[c]{0.33\textwidth}
  \centering
  \includegraphics[width=1\linewidth]{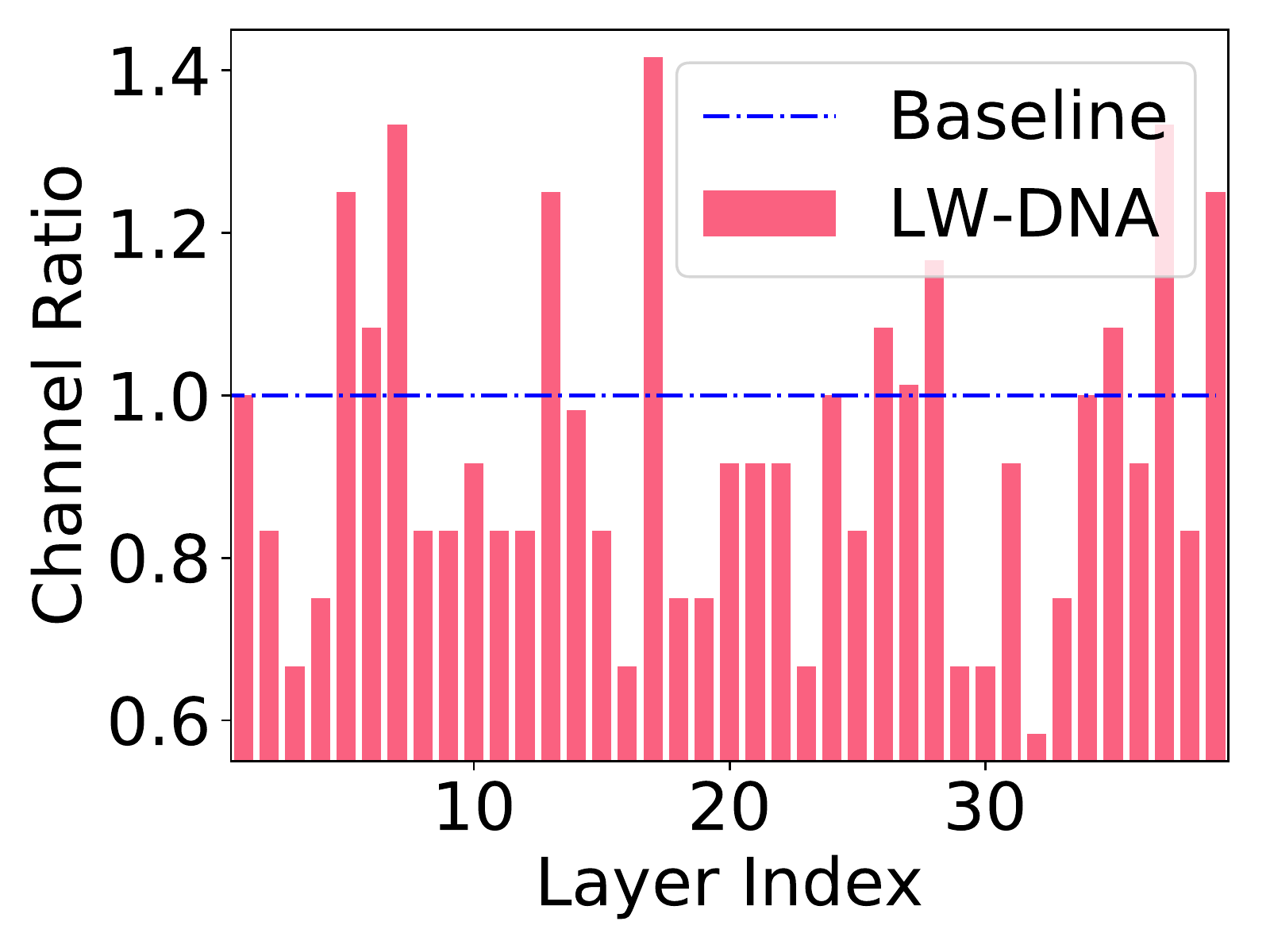}
  \subcaption{ DenseNet, CIFAR100.}
\end{minipage}%
\begin{minipage}[c]{0.33\textwidth}
  \centering
  \includegraphics[width=1\linewidth]{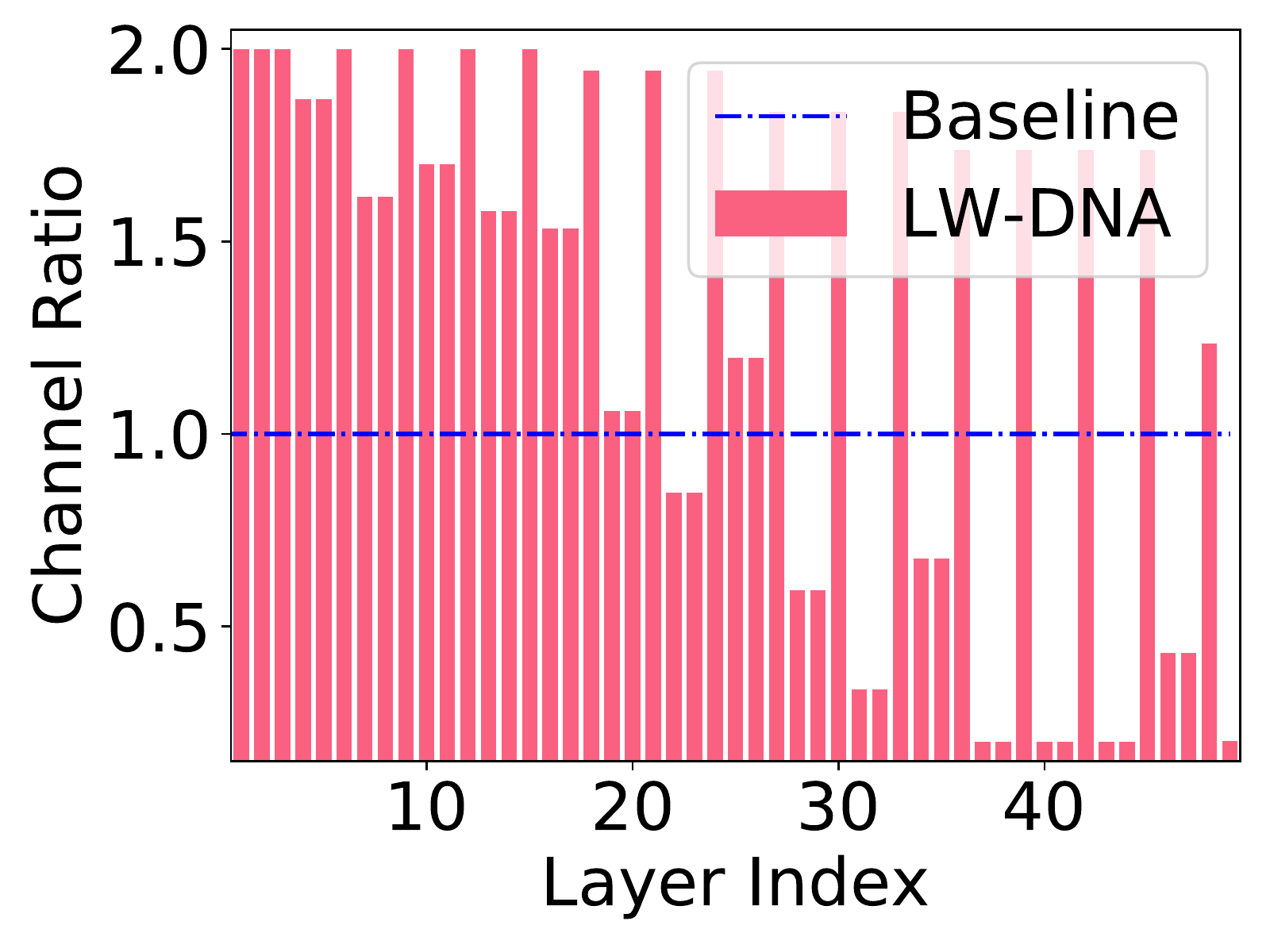}
  \subcaption{ EfficientNet, Tiny-ImageNet.}
\end{minipage}%
\begin{minipage}[c]{0.33\textwidth}
  \centering
  \includegraphics[width=1\linewidth]{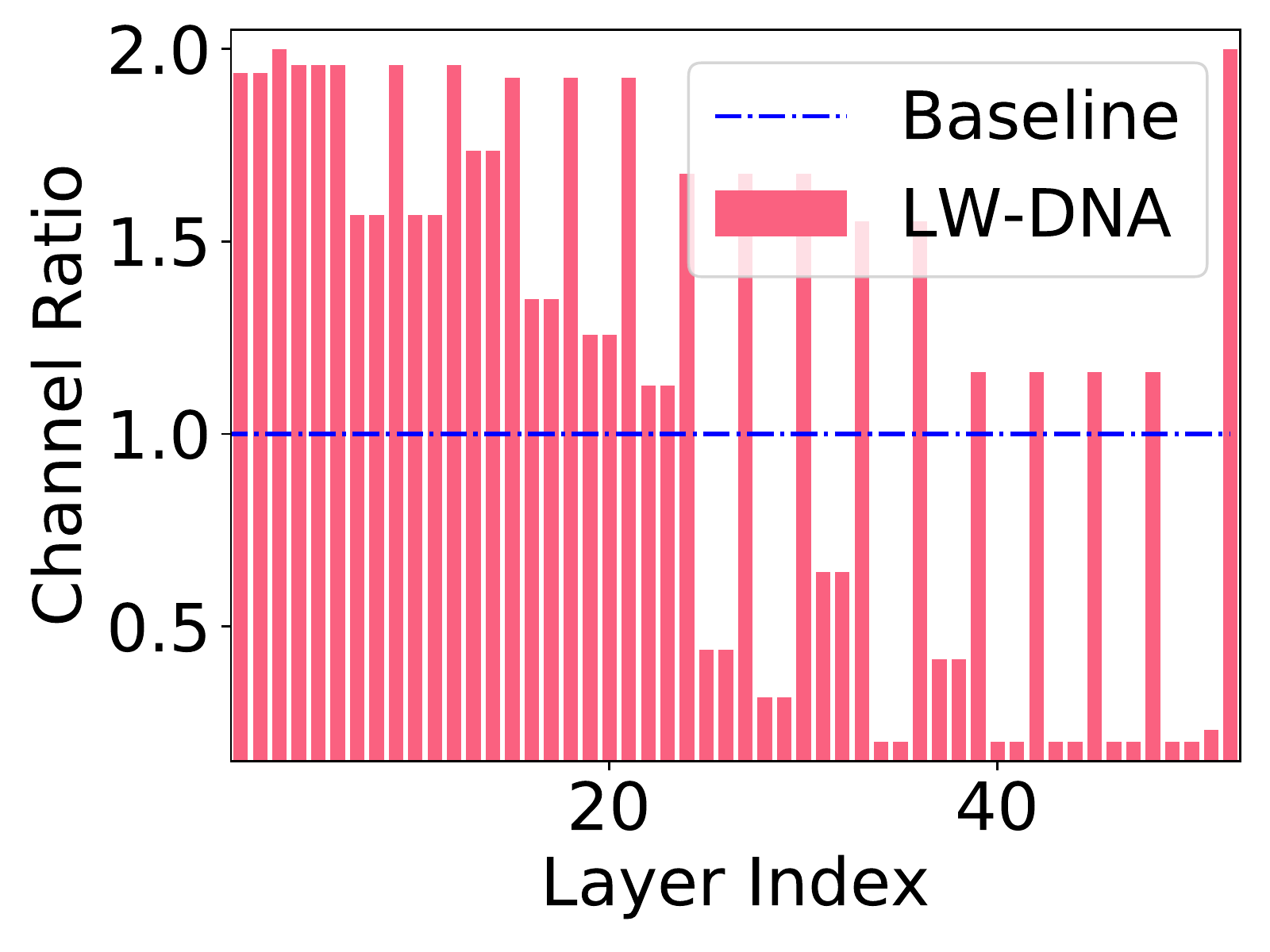}
  \subcaption{ MNASNet, Tiny-ImageNet.}
\end{minipage}%

\begin{minipage}[c]{0.33\textwidth}
  \centering
  \includegraphics[width=1\linewidth]{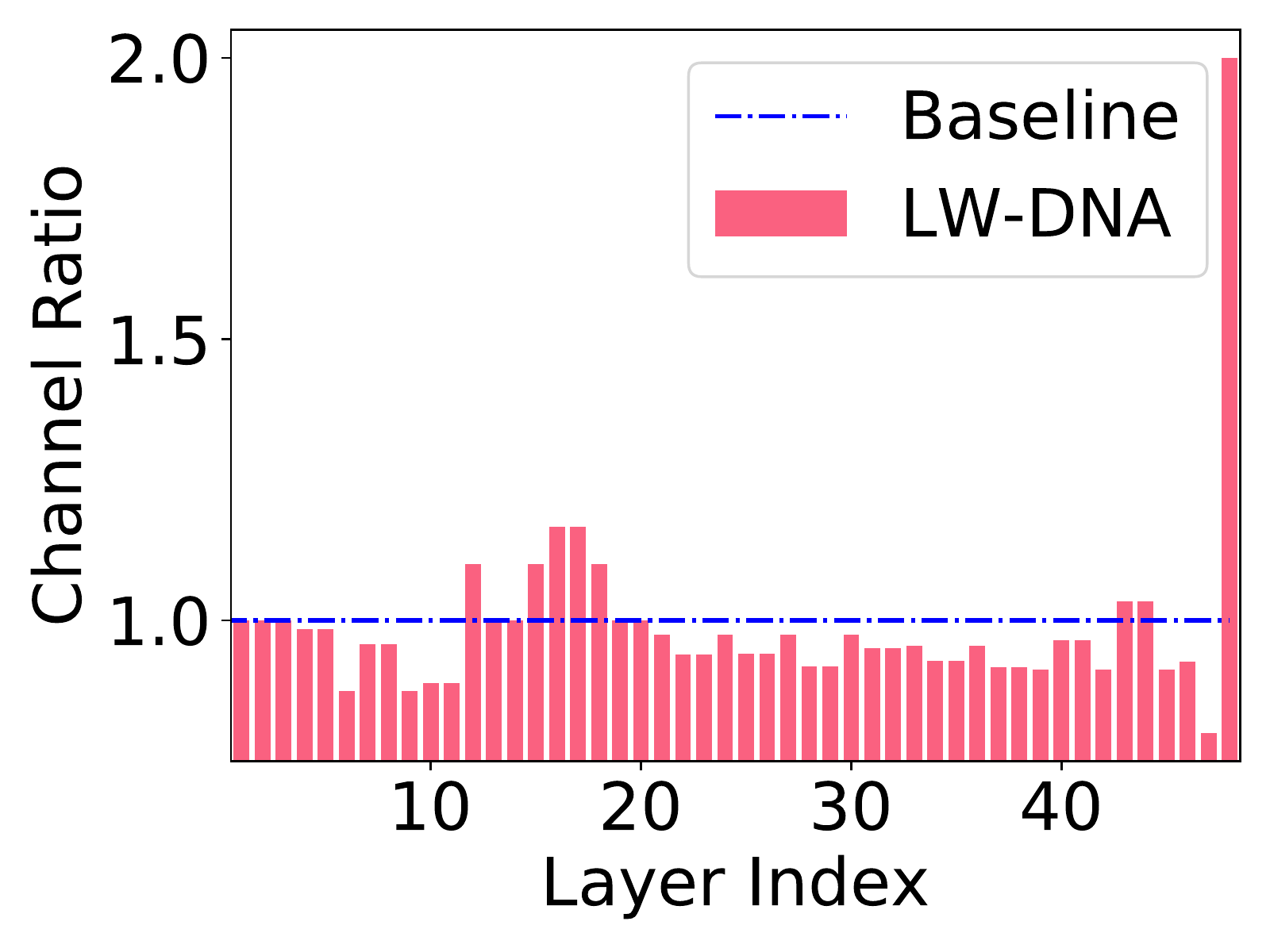}
  \subcaption{{MobileNetV3-large, Tiny-ImageNet.}}
\end{minipage}%
\begin{minipage}[c]{0.33\textwidth}
  \centering
  \includegraphics[width=1\linewidth]{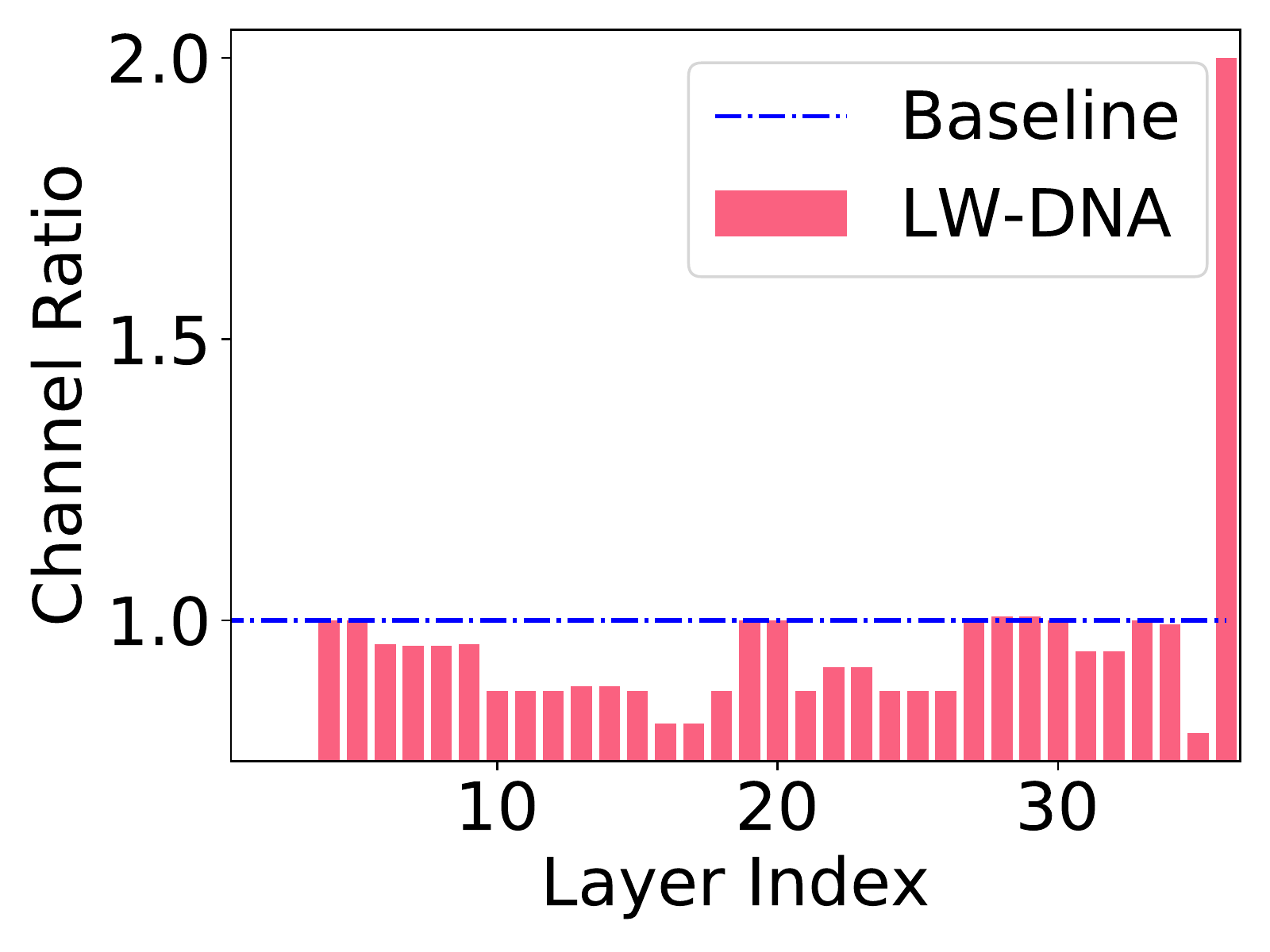}
  \subcaption{{MobileNetV3-small, Tiny-ImageNet.}}
\end{minipage}%
\begin{minipage}[c]{0.33\textwidth}
  \centering
  \includegraphics[width=1\linewidth]{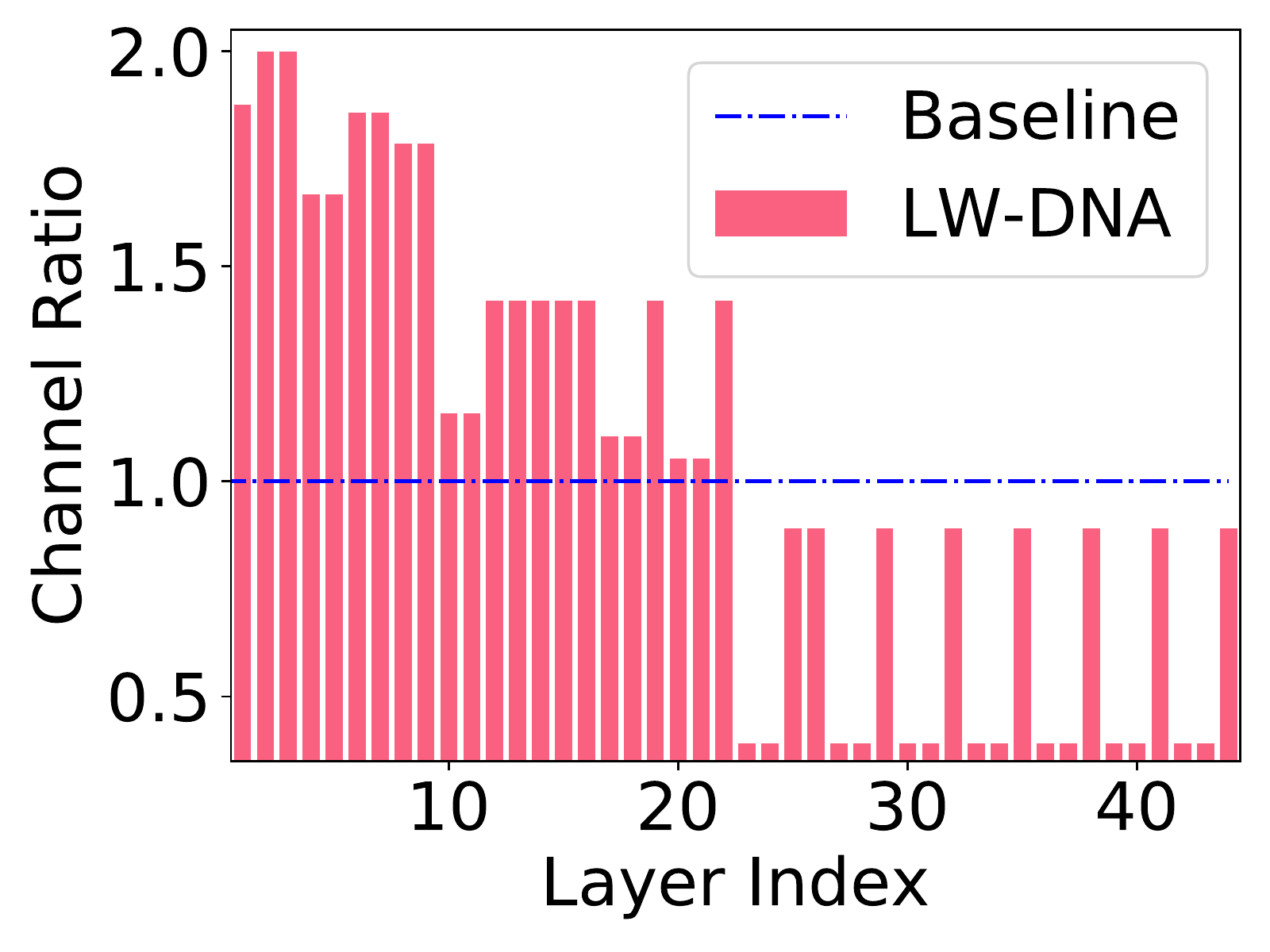}
  \subcaption{{RegNet 200MF, Tiny-ImageNet.}}
\end{minipage}%
\caption{Percentage of remaining output channels of LW-DNA models over the baseline network}
\label{fig:channel_percentage}
\end{figure*}

\begin{figure*}[!htb]\scriptsize
    \resizebox{\linewidth}{!}
    {
    \begin{tabular}{c@{\extracolsep{0.0em}}c@{\extracolsep{0.0em}}c@{\extracolsep{0.0em}}c@{\extracolsep{0.0em}}c@{\extracolsep{0.0em}}c}
		\includegraphics[width=0.255\textwidth]{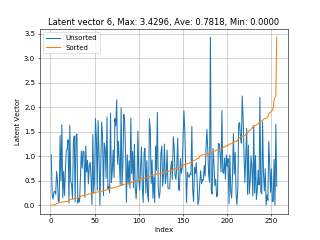}~
		&\includegraphics[width=0.255\textwidth]{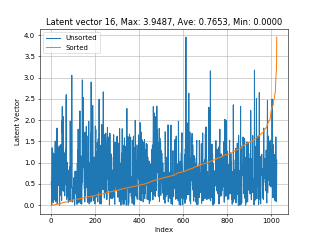}~
		&\includegraphics[width=0.255\textwidth]{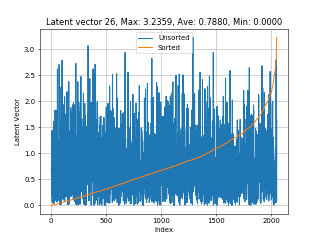} \\
		Layer 6, Epoch 1  &  Layer 16, Epoch 1  & Layer 26, Epoch 1 \\
		\includegraphics[width=0.255\textwidth]{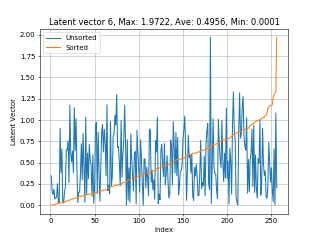}~
		&\includegraphics[width=0.255\textwidth]{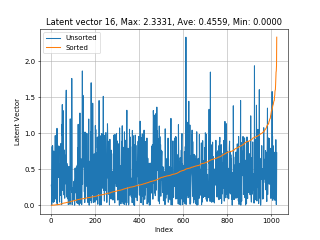}~
		&\includegraphics[width=0.255\textwidth]{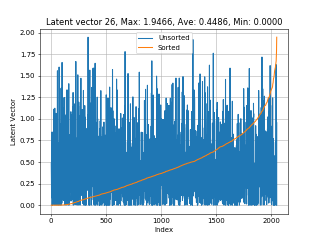} \\
		Layer 6, Epoch 4  &  Layer 16, Epoch 4  & Layer 26, Epoch 4 \\
		\includegraphics[width=0.255\textwidth]{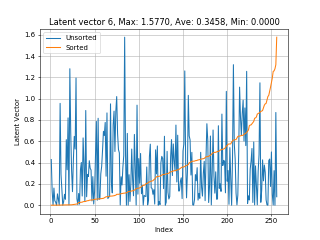}~
		&\includegraphics[width=0.255\textwidth]{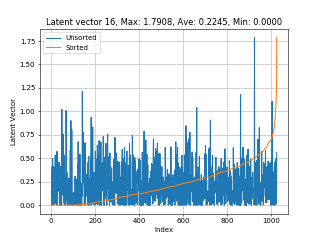}~
		&\includegraphics[width=0.255\textwidth]{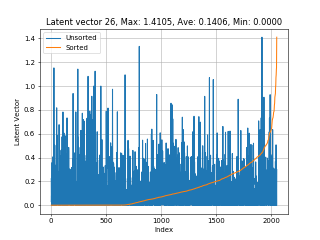} \\
		Layer 6, Epoch 10  &  Layer 16, Epoch 10  & Layer 26, Epoch 10 \\
		\includegraphics[width=0.255\textwidth]{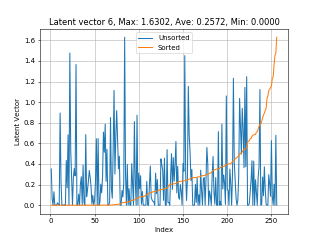}~
		&\includegraphics[width=0.255\textwidth]{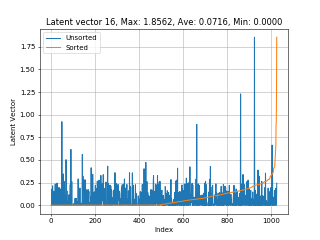}~
		&\includegraphics[width=0.255\textwidth]{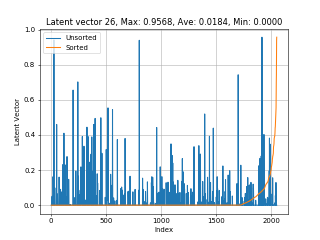} \\
		Layer 6, Epoch 17  &  Layer 16, Epoch 17  & Layer 26, Epoch 17 \\

	\end{tabular}
	}
	\caption{The distribution of the latent vectors in MobileNetV2 during the proximal gradient optimization of DHP.}
	\label{fig:latent_vector_distribution}
\end{figure*}

\textbf{Distribution of latent vectors}

The distribution of the latent vectors in MobileNetV2 during the DHP proximal gradient optimization is shown in Fig.~\ref{fig:latent_vector_distribution}. The distribution of the latent vectors at the end the optimization is related to the initial distribution to some extent. This phenomenon inspires us to pruning the latent vectors at initialization.

\end{document}